%% file: rbpn_cvpr2019.tex
\documentclass[10pt,twocolumn,letterpaper]{article}

\usepackage{cvpr}
\usepackage{times}
\usepackage{epsfig}
\usepackage{graphicx}
\usepackage{amsmath}
\usepackage{amssymb}
\usepackage{enumerate}
\usepackage{enumitem}
\usepackage{color}
\usepackage{comment}
\usepackage{multirow}
\usepackage{textcomp}

\usepackage[pagebackref=true,breaklinks=true,letterpaper=true,colorlinks,bookmarks=false]{hyperref}

\cvprfinalcopy 

\def\cvprPaperID{1927} 

\ifcvprfinal\fi
\begin{document}

\title{Recurrent Back-Projection Network for Video Super-Resolution}

\author{Muhammad Haris$^1$, Greg Shakhnarovich$^2$, and Norimichi Ukita$^1$\\
$^1$Toyota Technological Institute, Japan $^2$Toyota Technological Institute at Chicago, United States\\
{\tt\small \{mharis, ukita\}@toyota-ti.ac.jp, greg@ttic.edu}
}

\maketitle

\begin{abstract}
We proposed a novel architecture for the problem of video super-resolution. 
We integrate spatial and temporal contexts from continuous
video frames using a recurrent encoder-decoder module, that fuses
multi-frame information with the more traditional, single frame
super-resolution path for the target frame.
In contrast to most prior work where frames are pooled together by stacking or warping, our model, the Recurrent
Back-Projection Network (RBPN) treats each context frame as a separate
source of information. These sources are combined in an iterative
refinement framework inspired by the idea of back-projection in
multiple-image super-resolution. This is aided by explicitly representing estimated inter-frame
motion with respect to the target, rather than explicitly aligning
frames. We propose a new video super-resolution benchmark, allowing
evaluation at a larger scale and considering videos in different motion regimes.
Experimental results demonstrate that our RBPN is superior to existing methods on several datasets.

\end{abstract}

\input{introduction}
\input{related_work}
\input{method}

\input{experiments}

\input{conclusion}

\appendix
\section{Additional Experimental Results}
\subsection{Multiple scale factors}
To enrich the evaluation of RBPN, we provide the results for multiple
scaling factors (i.e., $2\times$ and $8\times$)
on Vimeo-90k~\cite{xue2017video}, SPMCS-32~\cite{tao2017detail}, and Vid4~\cite{liu2011bayesian} as shown in Table~\ref{tab:vimeo_results}.
Due to the limitation on other methods, the scores for other methods were copied from the respective publications.
RBPN is superior to existing methods on all test sets except the SSIM score on Vid4 in scaling factor $2\times$. However, note that the difference between the best score (i.e., VSR-DUF) and our score is only 0.001.

\begin{table*}[t!]
\small
\begin{center}
\begin{tabular}{*1c|*1l||*1c|*1c|*1c||*1c||*1c}
\hline
& &\multicolumn{3}{c||}{Vimeo-90k~\cite{xue2017video}} &SPMCS-32~\cite{tao2017detail} & Vid4~\cite{liu2011bayesian} \\ \cline{3-5}      
Scale&Algorithm  & Slow & Medium & Fast &&  \\
\hline \hline
&Bicubic						&34.47/0.939 &36.96/0.956 &40.18/0.969&32.39/0.919 &28.43/0.866\\
&DBPN~\cite{haris2018deep} &39.69/0.973 &42.33/0.979 &45.12/0.984&37.58/0.966 &32.30/0.934\\
2$\times$&VSR~\cite{kappeler2016video}*						&- &- &-&- &31.30/0.929\\
&DRDVSR~\cite{tao2017detail}*						&- &- &-&36.62/0.960 &-\\
&VSR-DUF~\cite{jo2018deep}*						&- &- &-&- &{\color{red}33.73/0.955}\\
&RBPN/6 &{\color{red}40.68/0.978} &{\color{red}43.65/0.985} &{\color{red}45.63/0.987}&{\color{red}39.26/0.977} &{\color{red}33.73}/0.954\\
\hline
\hline
&Bicubic 						&25.81/0.715 &27.23/0.766 &29.33/0.816&24.23/0.607 &21.36/0.465\\
8$\times$&DBPN~\cite{haris2018deep} &28.40/0.795 &30.28/0.837 &32.74/0.872&25.70/0.682 &22.39/0.547\\
&RBPN/6 &{\color{red}28.94/0.809} &{\color{red}31.35/0.858} &{\color{red}33.91/0.890}&{\color{red}26.31/0.708}&{\color{red}23.04/0.588}\\
\hline
\end{tabular}
\end{center}
\caption{Additional quantitative evaluation (PSNR/SSIM) of state-of-the-art SR algorithms. (*the values have been copied from the respective publications.)}
\label{tab:vimeo_results}
\end{table*}

\begin{table}[t!]
\small
\begin{center}
\begin{tabular}{*1l|*1c|*1c|*1c}
\hline
& RBPN/6-S & RBPN/6 & RBPN/6-L \\     
\hline
PSNR/SSIM &31.39/0.878 &{\color{red}31.64/0.883} & 31.58/0.882\\
\# of Parameter & 8,538$k$ & 12,771$k$ & 17,619$k$ \\
\hline
\end{tabular}
\end{center}
\caption{Network size analysis on SPMCS-32 (PSNR/SSIM). }
\label{tab:depth}
\end{table}

\begin{table}[t!]
\small
\begin{center}
\begin{tabular}{*1l|*1c|*1c}
\hline
 &\multicolumn{2}{c}{RBPN/6} \\ \cline{2-3}
& w/ & w/o \\     
\hline
PSNR/SSIM &31.57/0.882&{\color{red}31.64/0.883}\\
\hline
\end{tabular}
\end{center}
\caption{Residual analysis on SPMCS-32 (PSNR/SSIM). }
\label{tab:residual}
\end{table}

\subsection{Network size}
We observe the performance of RBPN on different network sizes.
The ``original'' RBPN use the same setup as in the main paper.
We use DBPN~\cite{haris2018deep} for $\texttt{Net}_{sisr}$, and Resnet~\cite{he2015deep} for $\texttt{Net}_{misr}$, $\texttt{Net}_{res}$, and $\texttt{Net}_{D}$.
For $\texttt{Net}_{sisr}$, we construct three stages using $8 \times
8$ kernel with stride = 4 and pad by 2 pixels.
For $\texttt{Net}_{misr}$, $\texttt{Net}_{res}$, and $\texttt{Net}_{D}$, we construct five blocks where each block consists of two convolutional layers with $3 \times 3$ kernel with stride = 1 and pad by 1 pixel.
The up-sampling layer in $\texttt{Net}_{misr}$ and down-sampling layer in $\texttt{Net}_{D}$ use $8 \times 8$ kernel with stride = 4 and pad by 2 pixels.
It also uses $c^l = 256, c^m = 256$, and $c^h = 64$.

RBPN-S uses $\texttt{Net}_{misr}$, $\texttt{Net}_{res}$, and $\texttt{Net}_{D}$ with three blocks, while RBPN-L uses deeper $\texttt{Net}_{sisr}$ with six stages.
The other setup remain the same.
Table~\ref{tab:depth} shows that RBPN/6 achieves the best performance.
The performance of RBPN/6 is reported in detail in the main paper.

\subsection{Residual Learning}
We also investigate the use of residual learning on RBPN. 
First, the target frame is interpolated using Bicubic, then RBPN only produces the residual image. 
Finally, the interpolated and residual images are combined to produce an SR image. 
Unfortunately, the current hyper-parameters show that residual learning is not effective to improve RBPN as shown in Table~\ref{tab:residual}.

\subsection{Complexity Analysis}
We report computational time, no. of parameter, and no. of FLOPS of
our proposal (and competition) in Table~\ref{tab:complexity}.

\begin{table}[t!]
\scriptsize
  \begin{center}
\begin{tabular}{*1l|*1c|*1c|*1c|*1c}
\hline
&RBPN/4-PF&RBPN/6-PF& VSR-DUF~\cite{jo2018deep} & DRDVSR~\cite{tao2017detail} \\     
\hline
Time ($s$) &{0.058} &0.141 &0.128 &{0.108}
\\
\# of param (M) &12.7 &12.7 & {6.8}* & {0.7}* \\
\# of FLOPS (G) & {1650} & {2475} & - & - \\
PSNR (dB)& {29.75} & {30.10} & 29.42 &28.82\\
\hline
\end{tabular}
\end{center}
\caption{Computational complexity on $4\times$ SR with input size $120\times160$. *Counted manually from model definitions described in the papers. }
\label{tab:complexity}\vspace{-1.5em}
\end{table}

\subsection{Additional Qualitative Results}
\begin{figure*}[t!]
  \begin{center}
    \begin{tabular}[c]{cccccc}
      \includegraphics[width=.155\textwidth]{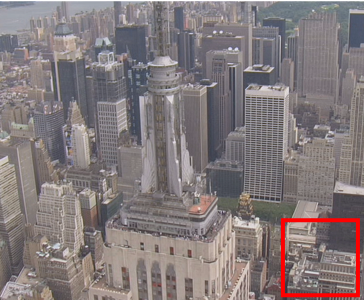}\hspace{-1em}\vspace{-0.3em}
      &
      \includegraphics[width=.15\textwidth]{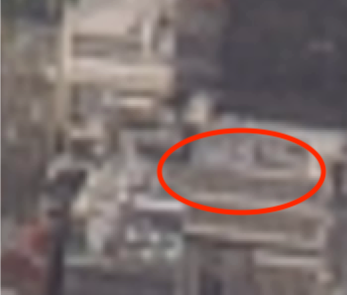}\hspace{-1em}
      &
      \includegraphics[width=.15\textwidth]{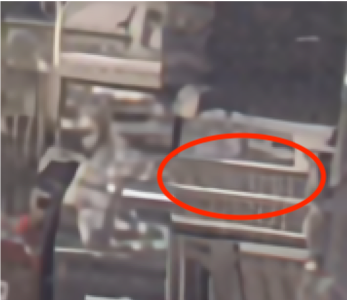}\hspace{-1em}
      &
      \includegraphics[width=.15\textwidth]{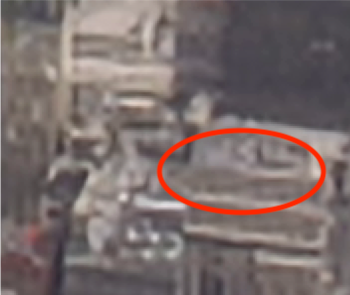}\hspace{-1em}
       &
       \includegraphics[width=.15\textwidth]{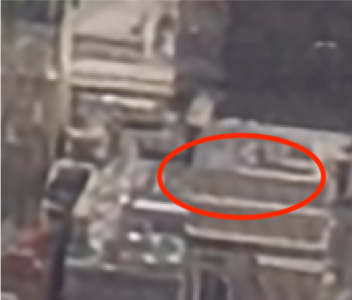}\hspace{-1em}
        &
        \includegraphics[width=.15\textwidth]{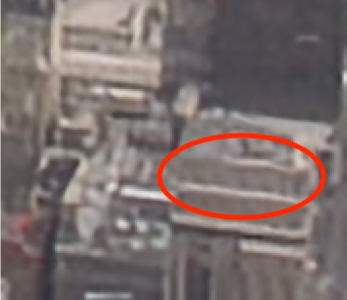}\hspace{-1em}\\
        ``City''&{\small (a) Bicubic\hspace{-1em}}
      &{\small (b) DBPN~\cite{haris2018deep}\hspace{-1em}}
      &{\small (c) VSR~\cite{kappeler2016video}\hspace{-1em}}
      &{\small (d) VESPCN~\cite{caballero2017real}\hspace{-1em}}
      &{\small (e) $B_{123}+T$~\cite{liu2017robust} \vspace{0.1em}}\\
        
      &
      \includegraphics[width=.15\textwidth]{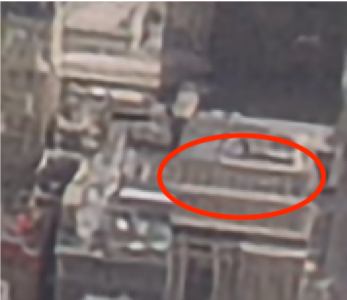}\hspace{-1em}\vspace{-0.3em}
      &
      \includegraphics[width=.15\textwidth]{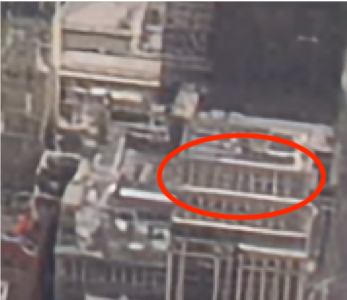}\hspace{-1em}
      &
      \includegraphics[width=.15\textwidth]{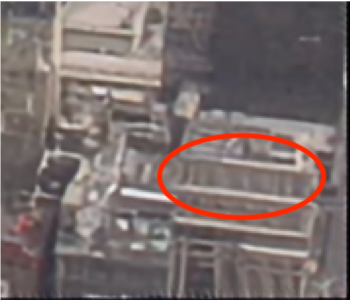}\hspace{-1em}
       &
       \includegraphics[width=.15\textwidth]{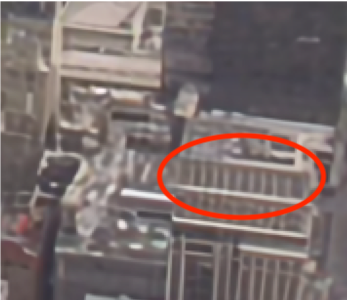}\hspace{-1em}
        &
        \includegraphics[width=.15\textwidth]{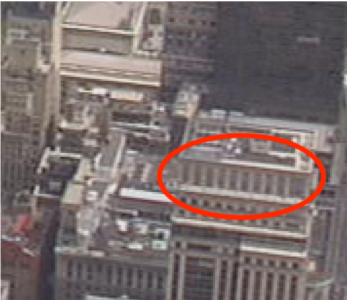}\hspace{-1em}\\
        &{\small (f) DRDVSR~\cite{tao2017detail}\hspace{-1em}}
      &{\small (g) FRVSR~\cite{sajjadi2018frame}\hspace{-1em}}
      &{\small (h) VSR-DUF~\cite{jo2018deep}\hspace{-1em}}
      &{\small (i) RBPN/6-PF\hspace{-1em}}
      &{\small (j) GT \vspace{0.1em}}\\

         \includegraphics[width=.16\textwidth]{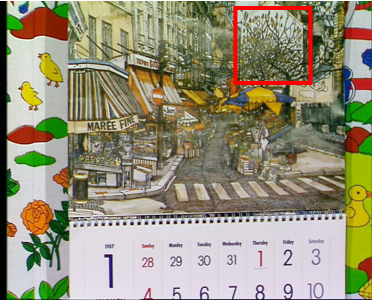}\hspace{-1em}\vspace{-0.3em}
      &
      \includegraphics[width=.15\textwidth]{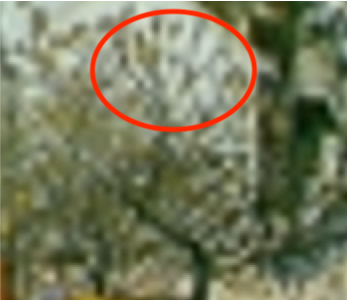}\hspace{-1em}
      &
      \includegraphics[width=.15\textwidth]{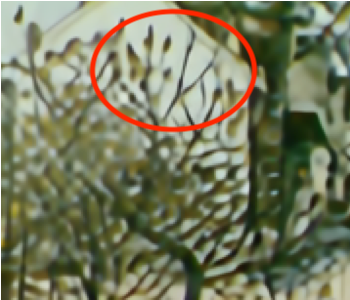}\hspace{-1em}
      &
      \includegraphics[width=.15\textwidth]{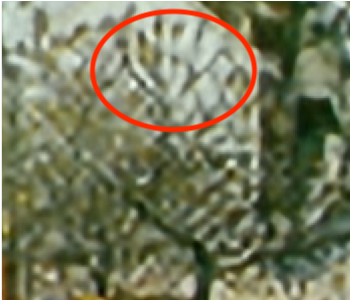}\hspace{-1em}
       &
       \includegraphics[width=.15\textwidth]{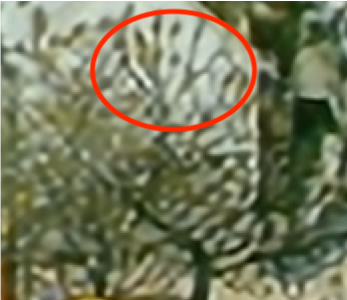}\hspace{-1em}
        &
        \includegraphics[width=.15\textwidth]{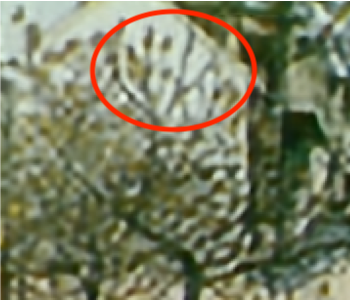}\hspace{-1em}\\
        ``Calendar''&{\small (a) Bicubic\hspace{-1em}}
      &{\small (b) DBPN~\cite{haris2018deep}\hspace{-1em}}
      &{\small (c) VSR~\cite{kappeler2016video}\hspace{-1em}}
      &{\small (d) VESPCN~\cite{caballero2017real}\hspace{-1em}}
      &{\small (e) $B_{123}+T$~\cite{liu2017robust} \vspace{0.1em}}\\
        
      &
      \includegraphics[width=.15\textwidth]{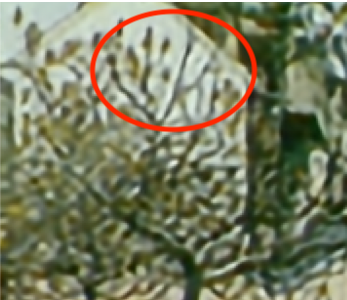}\hspace{-1em}\vspace{-0.3em}
      &
      \includegraphics[width=.15\textwidth]{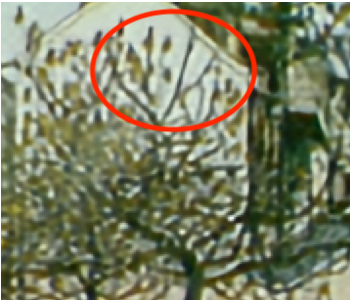}\hspace{-1em}
      &
      \includegraphics[width=.15\textwidth]{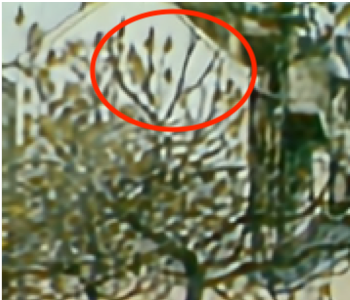}\hspace{-1em}
       &
       \includegraphics[width=.15\textwidth]{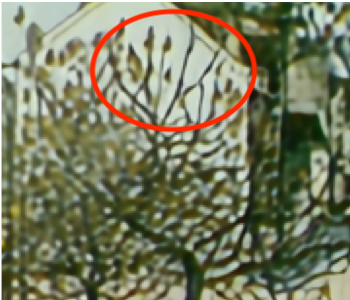}\hspace{-1em}
        &
        \includegraphics[width=.15\textwidth]{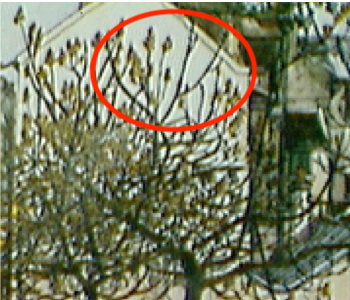}\hspace{-1em}\\
        
      &{\small (f) DRDVSR~\cite{tao2017detail}\hspace{-1em}}
      &{\small (g) FRVSR~\cite{sajjadi2018frame}\hspace{-1em}}
      &{\small (h) VSR-DUF~\cite{jo2018deep}\hspace{-1em}}
      &{\small (i) RBPN/6-PF\hspace{-1em}}
      &{\small (j) GT \vspace{0.2em}}\\
    \end{tabular}
    \caption{Visual results on Vid4 for $4\times$ scaling factor. }
    \label{fig:result_vid4}
  \end{center}
\end{figure*}

\begin{figure*}[t!]
  \begin{center}
    \begin{tabular}[c]{cccccc}
        \includegraphics[width=.2258\textwidth]{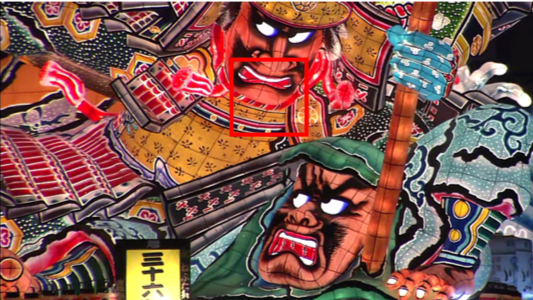}\hspace{-1em}
      &
      \includegraphics[width=.15\textwidth]{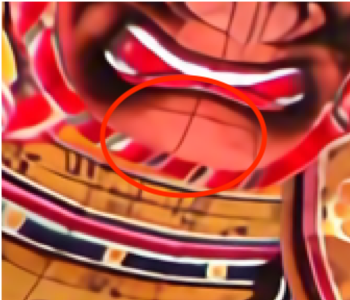}\hspace{-1em}
      &
      \includegraphics[width=.15\textwidth]{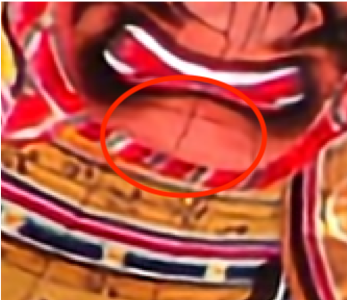}\hspace{-1em}
      &
      \includegraphics[width=.15\textwidth]{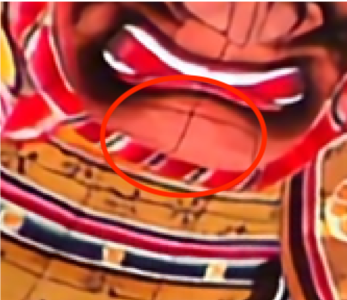}\hspace{-1em}
       &
       \includegraphics[width=.15\textwidth]{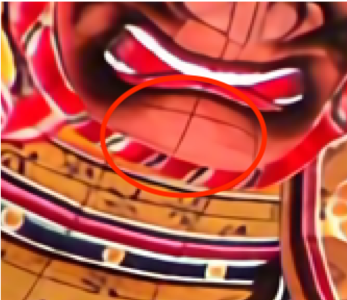}\hspace{-1em}
        &
        \includegraphics[width=.15\textwidth]{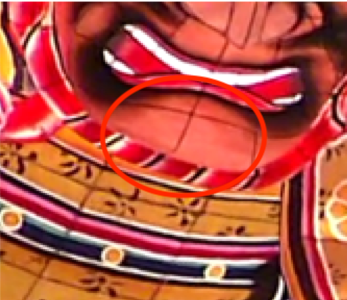}\hspace{-1em}\\
      &{\small (a) DBPN~\cite{haris2018deep}\hspace{-1em}}
      &{\small (b) DRDVSR~\cite{tao2017detail}\hspace{-1em}}
      &{\small (c) VSR-DUF~\cite{jo2018deep}\hspace{-1em}}
      &{\small (d) RBPN/6-PF\hspace{-1em}}
      &{\small (e) GT \vspace{0.2em}}\\
    \end{tabular}
    \caption{Visual results on SPMCS for $4\times$ scaling factor. }
    \label{fig:result_spmcs}
  \end{center}
\end{figure*}

\begin{figure*}
  \begin{center}
    \begin{tabular}[c]{cccccc}
      \includegraphics[width=.15\textwidth]{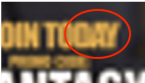}\hspace{-0.9em}
      &
      \includegraphics[width=.15\textwidth]{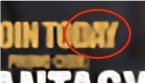}\hspace{-0.9em}
      &
      \includegraphics[width=.15\textwidth]{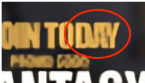}\hspace{-0.9em}
      &      
      \includegraphics[width=.15\textwidth]{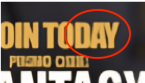}\hspace{-0.9em}
       &
       \includegraphics[width=.15\textwidth]{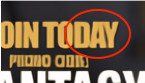}\hspace{-0.9em}
        &
        \includegraphics[width=.15\textwidth]{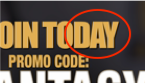}\hspace{-0.9em}\\
        
      \includegraphics[width=.15\textwidth]{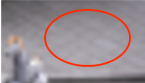}\hspace{-0.9em}
      &
      \includegraphics[width=.15\textwidth]{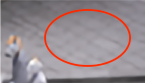}\hspace{-0.9em}
      &
      \includegraphics[width=.15\textwidth]{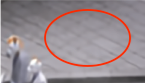}\hspace{-0.9em}
      &
      \includegraphics[width=.15\textwidth]{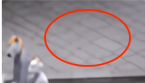}\hspace{-0.9em}
       &
       \includegraphics[width=.15\textwidth]{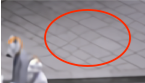}\hspace{-0.9em}
        &
        \includegraphics[width=.15\textwidth]{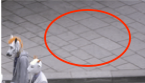}\hspace{-0.9em}\\
        
      \includegraphics[width=.15\textwidth]{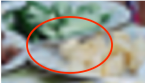}\hspace{-0.9em}
      &
      \includegraphics[width=.15\textwidth]{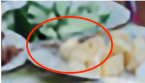}\hspace{-0.9em}
      &
      \includegraphics[width=.15\textwidth]{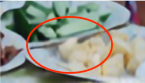}\hspace{-0.9em}
      &
      \includegraphics[width=.15\textwidth]{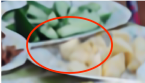}\hspace{-0.9em}
       &
       \includegraphics[width=.15\textwidth]{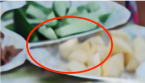}\hspace{-0.9em}
        &
        \includegraphics[width=.15\textwidth]{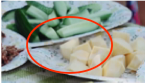}\hspace{-0.9em}\\

      \includegraphics[width=.15\textwidth]{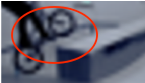}\hspace{-0.9em}
      &
      \includegraphics[width=.15\textwidth]{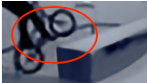}\hspace{-0.9em}
      &
      \includegraphics[width=.15\textwidth]{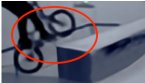}\hspace{-0.9em}
      &
      \includegraphics[width=.15\textwidth]{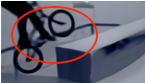}\hspace{-0.9em}
       &
       \includegraphics[width=.15\textwidth]{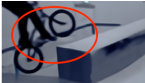}\hspace{-0.9em}
        &
        \includegraphics[width=.15\textwidth]{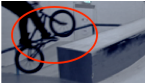}\hspace{-0.9em}\\
        
              \includegraphics[width=.15\textwidth]{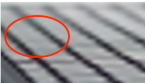}\hspace{-0.9em}
      &
      \includegraphics[width=.15\textwidth]{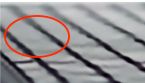}\hspace{-0.9em}
      &
      \includegraphics[width=.15\textwidth]{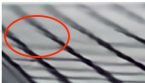}\hspace{-0.9em}
      &
      \includegraphics[width=.15\textwidth]{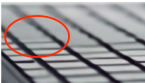}\hspace{-0.9em}
       &
       \includegraphics[width=.15\textwidth]{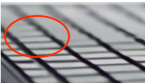}\hspace{-0.9em}
        &
        \includegraphics[width=.15\textwidth]{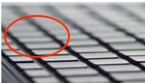}\hspace{-0.9em}\\
        
              \includegraphics[width=.15\textwidth]{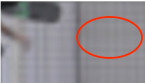}\hspace{-0.9em}
      &
      \includegraphics[width=.15\textwidth]{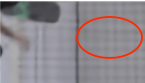}\hspace{-0.9em}
      &
      \includegraphics[width=.15\textwidth]{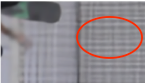}\hspace{-0.9em}
      &
      \includegraphics[width=.15\textwidth]{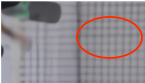}\hspace{-0.9em}
       &
       \includegraphics[width=.15\textwidth]{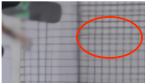}\hspace{-0.9em}
        &
        \includegraphics[width=.15\textwidth]{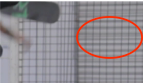}\hspace{-0.9em}\\
        
         \includegraphics[width=.15\textwidth]{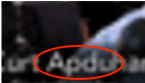}\hspace{-0.9em}
      &
      \includegraphics[width=.15\textwidth]{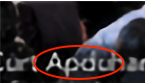}\hspace{-0.9em}
      &
      \includegraphics[width=.15\textwidth]{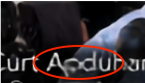}\hspace{-0.9em}
      &
      \includegraphics[width=.15\textwidth]{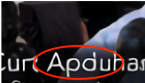}\hspace{-0.9em}
       &
       \includegraphics[width=.15\textwidth]{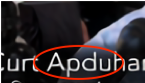}\hspace{-0.9em}
        &
        \includegraphics[width=.15\textwidth]{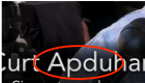}\hspace{-0.9em}\\

         {\small (a) Bicubic\hspace{-0.9em}}
      &{\small (b) TOFlow~\cite{xue2017video}\hspace{-0.9em}}
      &{\small (c) VSR-DUF~\cite{jo2018deep}\hspace{-0.9em}}
      &{\small (d) RBPN/3-P\hspace{-0.9em}}
      &{\small (e) RBPN/6-PF\hspace{-0.9em}}
      &{\small (f) GT \vspace{0.3em}}\\
    \end{tabular}
    \caption{Visual results on Vimeo-90k for $4\times$ scaling factor. }
    \label{fig:result_vimeo}
  \end{center}
\end{figure*}

Here, we show additional results on several test sets and scaling factors.
Figures~\ref{fig:result_vid4},~\ref{fig:result_spmcs}, and~\ref{fig:result_vimeo} show qualitative results for the $4\times$ scaling factor on Vid4~\cite{liu2011bayesian}, SPMCS-32~\cite{tao2017detail}, and Vimeo-90k~\cite{xue2017video}, respectively. 
RBPN/6-PF obtains reconstruction that appears most similar to the GT,
more pleasing and sharper than reconstructions with other methods. We
have highlighted regions in which this is particularly notable.

\begin{figure*}[t!]
  \begin{center}
    \begin{tabular}[c]{ccccc}
      \includegraphics[width=.15\textwidth]{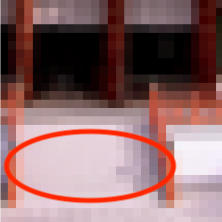}\hspace{-0.9em}
      &
      \includegraphics[width=.15\textwidth]{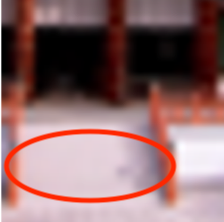}\hspace{-0.9em}
      &
      \includegraphics[width=.15\textwidth]{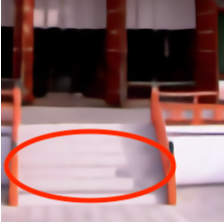}\hspace{-0.9em}
      &      
      \includegraphics[width=.15\textwidth]{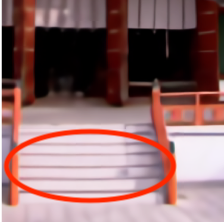}\hspace{-0.9em}
       &
       \includegraphics[width=.15\textwidth]{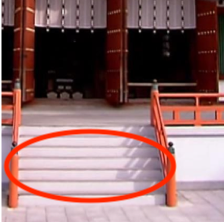}\hspace{-0.9em}\\  
       
        \includegraphics[width=.15\textwidth]{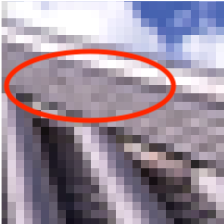}\hspace{-0.9em}
      &
      \includegraphics[width=.15\textwidth]{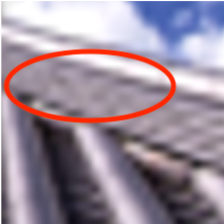}\hspace{-0.9em}
      &
      \includegraphics[width=.15\textwidth]{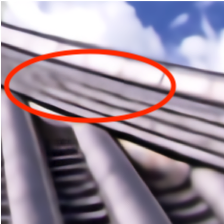}\hspace{-0.9em}
      &      
      \includegraphics[width=.15\textwidth]{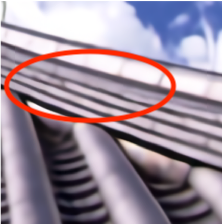}\hspace{-0.9em}
       &
       \includegraphics[width=.15\textwidth]{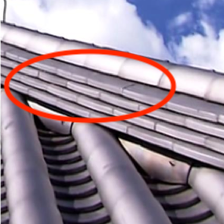}\hspace{-0.9em}\\  
       
        \includegraphics[width=.15\textwidth]{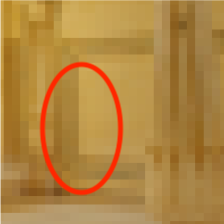}\hspace{-0.9em}
      &
      \includegraphics[width=.15\textwidth]{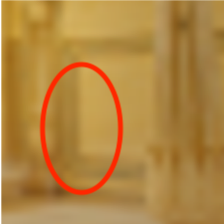}\hspace{-0.9em}
      &
      \includegraphics[width=.15\textwidth]{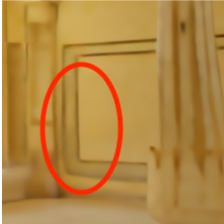}\hspace{-0.9em}
      &      
      \includegraphics[width=.15\textwidth]{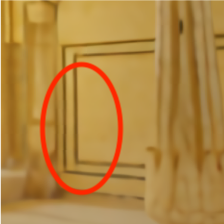}\hspace{-0.9em}
       &
       \includegraphics[width=.15\textwidth]{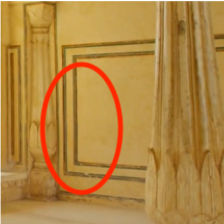}\hspace{-0.9em}\\  
       
       
        \includegraphics[width=.15\textwidth]{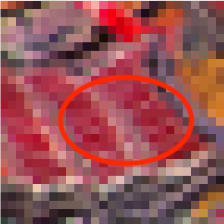}\hspace{-0.9em}
      &
      \includegraphics[width=.15\textwidth]{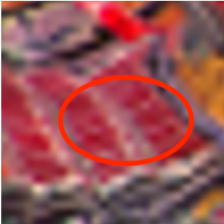}\hspace{-0.9em}
      &
      \includegraphics[width=.15\textwidth]{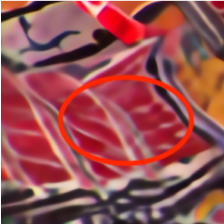}\hspace{-0.9em}
      &      
      \includegraphics[width=.15\textwidth]{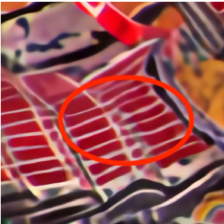}\hspace{-0.9em}
       &
       \includegraphics[width=.15\textwidth]{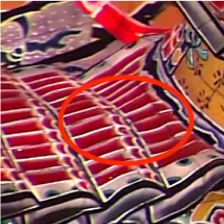}\hspace{-0.9em}\\    
       
        \includegraphics[width=.15\textwidth]{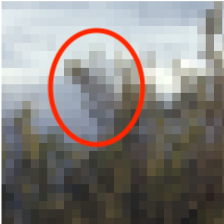}\hspace{-0.9em}
      &
      \includegraphics[width=.15\textwidth]{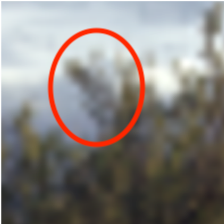}\hspace{-0.9em}
      &
      \includegraphics[width=.15\textwidth]{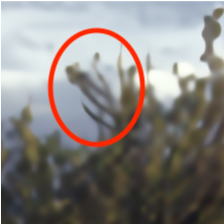}\hspace{-0.9em}
      &      
      \includegraphics[width=.15\textwidth]{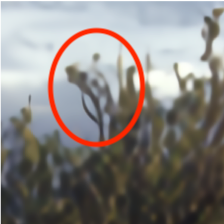}\hspace{-0.9em}
       &
       \includegraphics[width=.15\textwidth]{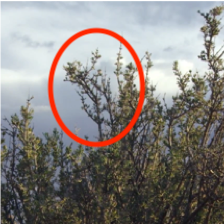}\hspace{-0.9em}\\  
       
        \includegraphics[width=.15\textwidth]{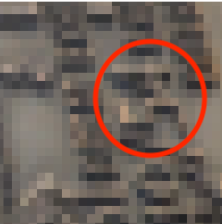}\hspace{-0.9em}
      &
      \includegraphics[width=.15\textwidth]{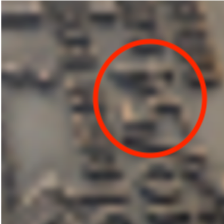}\hspace{-0.9em}
      &
      \includegraphics[width=.15\textwidth]{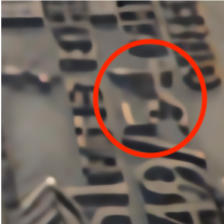}\hspace{-0.9em}
      &      
      \includegraphics[width=.15\textwidth]{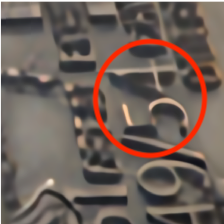}\hspace{-0.9em}
       &
       \includegraphics[width=.15\textwidth]{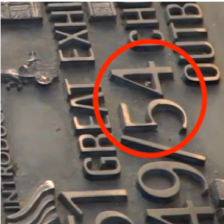}\hspace{-0.9em}\\  
       
        \includegraphics[width=.15\textwidth]{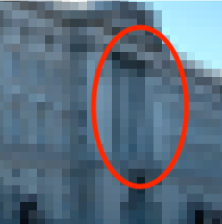}\hspace{-0.9em}
      &
      \includegraphics[width=.15\textwidth]{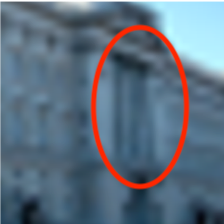}\hspace{-0.9em}
      &
      \includegraphics[width=.15\textwidth]{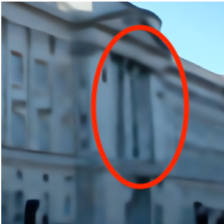}\hspace{-0.9em}
      &      
      \includegraphics[width=.15\textwidth]{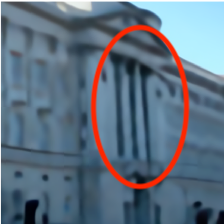}\hspace{-0.9em}
       &
       \includegraphics[width=.15\textwidth]{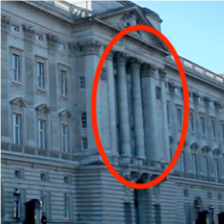}\hspace{-0.9em}\\  

         {\small (a) LR\hspace{-0.9em}}
      &{\small (b) Bicubic\hspace{-0.9em}}
      &{\small (c) DBPN~\cite{haris2018deep}\hspace{-0.9em}}
      &{\small (d) RBPN/6\hspace{-0.9em}}
      &{\small (e) GT \vspace{0.3em}}\\
    \end{tabular}
    \caption{Visual results on SPMCS for $8\times$ scaling factor.}
    \label{fig:result_spmcs_8x}
  \end{center}
\end{figure*}

We also provide the results on a larger scaling factor (i.e., $8\times$) in Fig.~\ref{fig:result_spmcs_8x}.
However, no results were provided by other methods on $8\times$, so we only compare ours with DBPN and Bicubic.
It shows that RBPN/6 successfully generates the best results.

{\small
\bibliographystyle{ieee}
\bibliography{egbib}
}

\end{document}

%% file: introduction.tex
\section{Introduction}
\label{intro}

The goal of super-resolution (SR) is to enhance a low-resolution (LR) image
to higher resolution (HR) by filling missing fine
details in the LR image. 
This field can be divided into Single-Image SR
(SISR)~\cite{dong2016image,haris2018deep,Haris17,Kim_2016_VDSR,LapSRN,Tai-DRRN-2017},
Multi-Image SR (MISR)~\cite{faramarzi2013unified,garcia2012super}, and
Video SR (VSR)~\cite{caballero2017real,tao2017detail,
  sajjadi2018frame, jo2018deep, huang2015bidirectional,
  liu2017robust}, the focus of this paper.

Consider a sequence of LR video frames
$I_{t-n},\ldots,I_{t},\ldots,I_{t+n}$,
where we super-resolve a \emph{target frame}, $I_t$.
While $I_{t}$ can be super-resolved independently of other frames as
SISR, this is wasteful of missing details available from the other
frames.
In MISR, the missing details available from
the other frames are fused for super-resolving $I_{t}$.
For extracting these missing details, all frames must be spatially
aligned explicitly or implicitly.
By separating differences between the aligned frames from missing
details observed only in
one or some of the frames, the missing details are extracted.
This alignment is required to be
very precise (e.g., sub-pixel accuracy) for SR.
In MISR, however, the frames are aligned independently with no cue
given by temporal smoothness, resulting in difficulty in the precise
alignment.
Yet another approach is to align the frames in temporal smooth order
as VSR.

In recent VSR methods using convolutional networks,
the frames are concatenated~\cite{liao2015video,jo2018deep}
or fed into recurrent networks (RNNs)~\cite{huang2015bidirectional})
in temporal order; no explicit alignment is performed.
The frames can be also aligned explicitly,
using motion cues between temporal frames
with the alignment modules~\cite{liu2017robust,caballero2017real,tao2017detail,sajjadi2018frame}.
These latter methods generally produce results superior to
those with no explicit spatial
alignment~\cite{liao2015video,huang2015bidirectional}.
Nonetheless, these VSR methods suffer from a number of problems.
%
In the frame-concatenation approach \cite{caballero2017real, jo2018deep, liu2017robust},
many frames are processed simultaneously in the network, resulting in
difficulty in training the network.
In RNNs~\cite{tao2017detail, sajjadi2018frame,
  huang2015bidirectional}, it is not easy to jointly model subtle and
significant changes (e.g., slow and quick motions of foreground
objects) observed in all frames of a video even by those designed for
maintaining
long-term temporal dependencies such as
LSTMs~\cite{gers1999learning}.

\begin{figure*}[!t]
  \begin{center}
    \begin{tabular}[c]{ccccc}
      \includegraphics[height=.14\textheight]{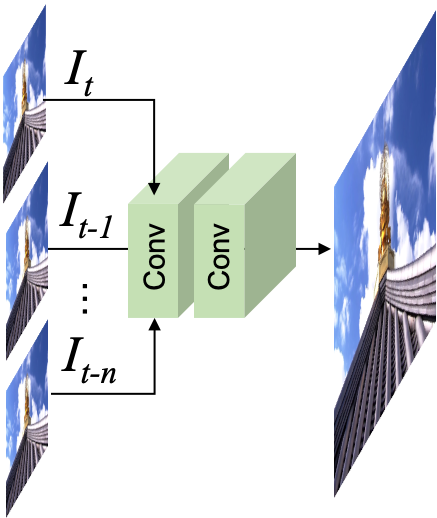} 
      &
        \includegraphics[height=.14\textheight]{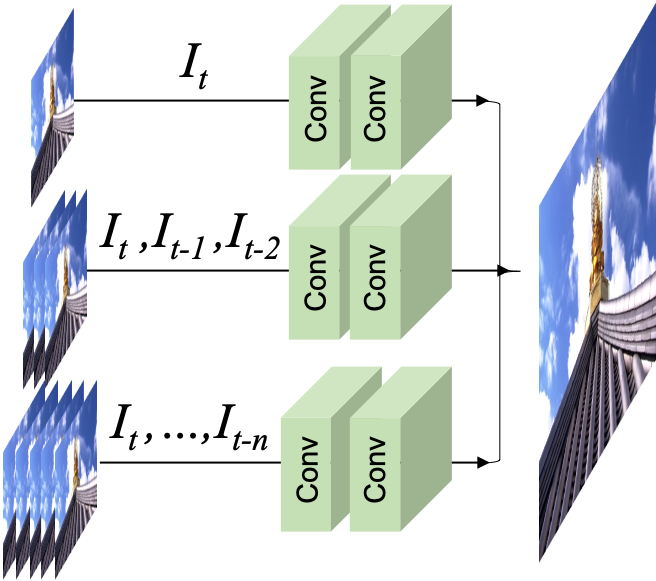} 
      &
        \includegraphics[height=.14\textheight]{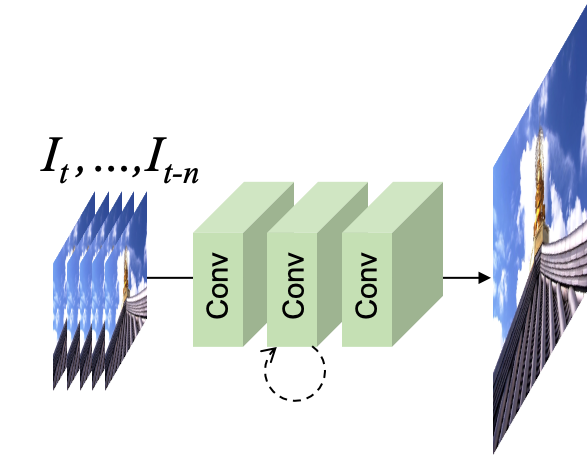} 
      &
        \includegraphics[height=.14\textheight]{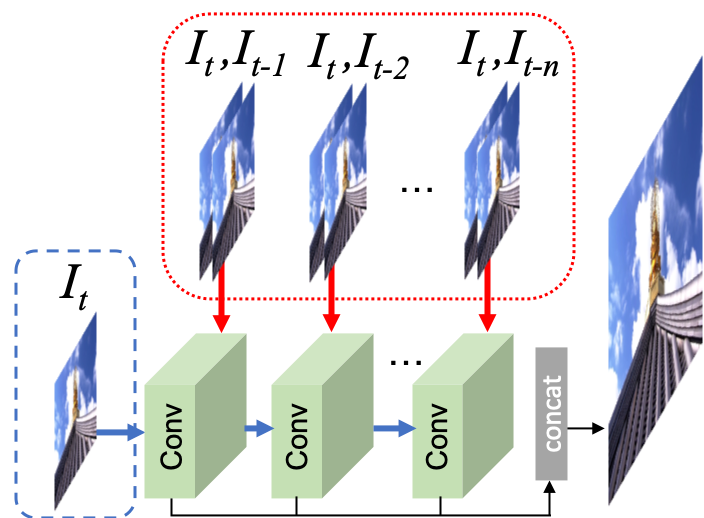}\\
        
      {\small (a) Temporal concatenation}
      &{\small (b) Temporal aggregation}
      &{\small (c) RNNs}
      &{\small (d) Our RBPN}\\
    \end{tabular}\vspace{0.2em}
    \caption{Comparison of Deep VSRs. (a) Input
      frames are concatenated to preserve temporal
      information~\cite{kappeler2016video,caballero2017real,jo2018deep,liao2015video}.
      (b) Temporal aggregation improves (a) to preserve
      multiple motion regimes~\cite{liu2017robust}.
      (c) RNNs take a sequence of input frames
      to produce one SR image at a target frame,
      $I_{t}$~\cite{huang2015bidirectional, tao2017detail,
        sajjadi2018frame}.
      (d) Our
      recurrent back-projection network accepts $I_{t}$, which is
      enclosed by a blue dashed line, as well as a
      set of residual features computed from a pairing $I_{t}$ with
      other frames (i.e., $I_{t-k}$ for
      $k \in \{ 1,\cdots,n \}$), as enclosed by a red dotted line,
      while previous approaches using RNNs shown in (c)
      feed all temporal frames one by one along a single path.
%
%
      Residual features computed from the pairs of $(I_{t},I_{t-k})$
      (MISR path - {\color{red}the vertical red arrows}) are fused with features extracted
      from variants of $I_{t}$  (SISR path - {\color{blue}the horizontal blue arrows}) through RNN.
%
%
    }
    \label{fig:deep_vsr}\vspace{-1em}
  \end{center}
\end{figure*}

Our method proposed in this paper is inspired by
``back-projection'' originally introduced
in~\cite{irani1991improving,irani93} for MISR.
Back-projection iteratively calculates residual images as
reconstruction error between a target image and a set of its
corresponding images. The residuals are back-projected to the target
image for improving its resolution.
The multiple residuals can represent subtle and significant
differences between the target frame and other frames independently.
Recently, Deep Back-Projection Networks (DBPN)~\cite{haris2018deep} extended back-projection to Deep
SISR under the assumption that only one LR image is
given for the target image.
In that scenario, DBPN produces a high-resolution feature map, iteratively refined 
through multiple up- and down-sampling layers.
Our method, Recurrent Back-Projection Networks (RBPN), integrates
the benefits of the original, MISR back projection and DBPN,
for VSR. Here we use other video frames as
corresponding LR images for the original MISR back-projection. In addition,
we use the idea of iteratively refining HR feature maps representing
missing details by 
up- and down-sampling processes to further improve the quality of SR.

Our contributions include the following key innovations.

\noindent\textbf{Integrating SISR and MISR in a unified VSR framework:}
SISR and MISR extract missing details from different sources.
Iterative SISR~\cite{haris2018deep} extracts various feature maps
representing the details of a target frame. MISR provides multiple
sets of feature maps from other frames.
These different sources are iteratively
updated in temporal order 
through RNN for VSR.

\noindent\textbf{Back-projection modules for RBPN:} We develop a
recurrent encoder-decoder mechanism for incorporating details
extracted in SISR and MISR paths through the back-projection.
%
While the SISR path accepts only $I_{t}$, the MISR path also accepts
$I_{t-k}$ where $k \in [n]$. A gap between
$I_{t}$ and $I_{t-k}$ is larger than the one in other VSRs using RNN
(i.e., gap only between $I_{t}$ and $I_{t-1}$).
Here, the network is able to understand this large gap since each context is calculated separately, rather than jointly as
in previous work, this separate context plays an important role in RBPN.


\noindent\textbf{Extended evaluation protocol:} We report extensive
experiments to evaluate VSR.
In addition to previously-standard datasets,
Vid4~\cite{liu2011bayesian} and SPMCS~\cite{tao2017detail}, that lack
significant motions, a dataset containing various types of motion
(Vimeo-90k~\cite{xue2017video}) is used in our evaluation.
This allows us to conduct a more detailed evaluation of strengths and
weaknesses of VSR methods, depending on the type of
the input video.

%% file: related_work.tex
\section{Related Work}
\label{sec:related}
\begin{figure*}[t!]
\centering
\includegraphics[width=13cm]{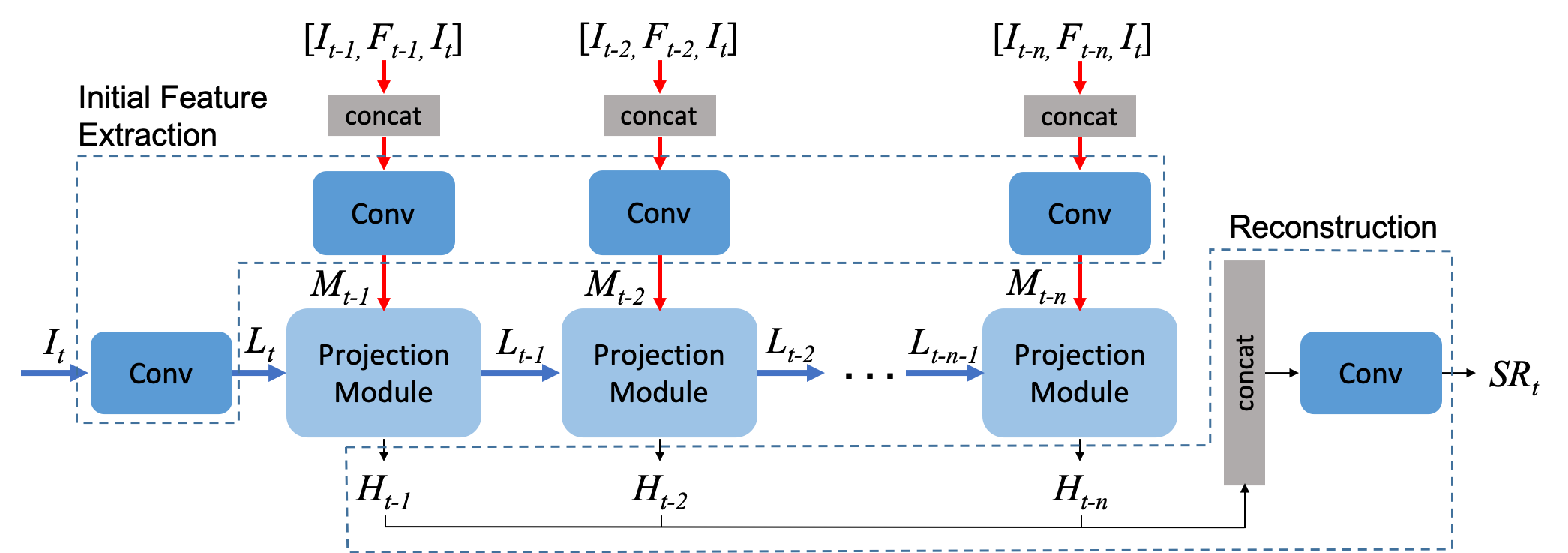}
\caption{Overview of RBPN. The network has two approaches. The horizontal blue line enlarges $I_t$ using SISR. The vertical red line is based on MISR to compute the residual features from a pair of
$I_t$ to neighbor frames ($I_{t-1}, ..., I_{t-n}$) and the precomputed dense motion flow maps ($F_{t-1}, ..., F_{t-n}$). Each step is connected to add the temporal connection.
On each projection step, RBPN observes the missing details on $I_t$ and extract the residual features from each neighbor frame to recover the missing details.}
\label{figure:proposed_network}
\end{figure*}

While SR has an extensive history, our discussion in this section focuses on \emph{deep SR} -- SR methods
that involve deep neural network components, trained end-to-end.

\subsection{Deep Image Super-resolution }
Deep SISR is first introduced by SRCNN~\cite{dong2016image} that requires a predefined upsampling operator.
Further improvements include better up-sampling layers~\cite{shi2016real}, residual learning~\cite{Kim_2016_VDSR,Tai-DRRN-2017}, back-projection~\cite{haris2018deep}, recursive layers~\cite{kim2016deeply}, and progressive upsampling~\cite{LapSRN}.
See 
NTIRE2018~\cite{timofte2018ntire} and PIRM2018~\cite{pirm2018} for
comprehensive comparison.

\subsection{Recurrent Networks}

Recurrent neural networks (RNNs) deal with sequential inputs and/or outputs, and
have been employed for video
captioning~\cite{johnson2016densecap,mao2014deep,yu2016video}, video
summarization~\cite{donahue2015long,venugopalan2014translating}, and
VSR~\cite{tao2017detail,huang2015bidirectional,sajjadi2018frame}. Two
types of RNN have been used for VSR. A many-to-one architecture is used
in~\cite{tao2017detail,huang2015bidirectional} where a sequence of
frames is mapped to a single target HR frame. A synchronous many-to-many
RNN has recently been used in VSR by~\cite{sajjadi2018frame}, to map a
sequence of LR frames to a sequence of HR frames.
%

\subsection{Deep Video Super-resolution}

Deep VSR can be primarily divided into three types based on the
approach to preserving temporal information as shown in
Fig.~\ref{fig:deep_vsr} (a), (b), and (c).

\noindent(a) \textbf{Temporal Concatenation}. The most popular approach to retain temporal information in VSR is by concatenating the frames as in~\cite{kappeler2016video,caballero2017real,jo2018deep,liao2015video}. 
This approach can be seen as an extension of SISR to accept multiple input images. 
VSR-DUF~\cite{jo2018deep} proposed a mechanism to construct up-sampling filters and residual images. 
However, this approach fails to represent the multiple motion regimes on a sequence because input frames are concatenated together.

\noindent(b) \textbf{Temporal Aggregation.} To address the dynamic motion problem in VSR, \cite{liu2017robust}~proposed multiple SR inferences which work on different motion regimes. 
The final layer aggregates the outputs of all branches to construct  SR frame.
However, this approach basically still concatenates many input frames, resulting in difficulty in global optimization.


\noindent(c) \textbf{RNNs.} This approach is first
proposed by~\cite{huang2015bidirectional} using bidirectional
RNNs. However, the network has a small network capacity
and has no frame alignment step. Further improvement is proposed
by~\cite{tao2017detail} using a motion compensation module and a convLSTM layer~\cite{xingjian2015convolutional}. 
Recently, \cite{sajjadi2018frame} proposed an efficient many-to-many 
RNN that uses the previous HR estimate to super-resolve the next frames. 
While recurrent feedback connections utilize temporal smoothness
between neighbor frames in a video for improving the performance,
it is not easy to jointly model subtle and significant changes
observed in all frames.


%% file: method.tex
\section{Recurrent Back-Projection Networks}
\label{sec:rbpn} 
\subsection{Network Architecture}
Our proposed network is illustrated in Fig.~\ref{figure:proposed_network}.
Let $I$ be LR frame with size of $(M^l \times N^l)$. The input is
sequence of $n+1$ LR frames $\{I_{t-n},\ldots,I_{t-1},I_{t}\}$ where
$I_t$ is the target frame. The goal of VSR is to output HR version of
$I_t$, denoted by $\texttt{SR}_t$ with size of $(M^h \times N^h)$
where $M^l < M^h$ and $N^l < N^h$. The operation of RBPN can be
divided into three stages: initial feature extraction, multiple
projections, and reconstruction. Note that we train the entire
network jointly, end-to-end.

\noindent\textbf{Initial feature extraction}. 
Before entering projection modules, $I_t$ is mapped to LR feature
tensor $L_t$. 
For each neighbor frame among $I_{t-k}$, $k\in[n]$,
we concatenate the precomputed dense motion flow
map $F_{t-k}$ (describing a 2D vector per pixel) between $I_{t-k}$ and $I_t$ with
the target frame $I_t$ and $I_{t-k}$.
The motion flow map encourages the projection module to extract
missing details between a pair of $I_t$ and $I_{t-k}$.
This stacked 8-channel ``image'' is mapped to a \emph{neighbor
  feature} tensor $M_{t-k}$ .

\noindent\textbf{Multiple Projections}. 
Here, we extract the missing details in the target frame by integrating SISR and MISR paths, then produce refined HR feature tensor.
This stage receives $L_{t-k-1}$ and $M_{t-k}$, and outputs HR feature tensor $H_{t-k}$. 

\noindent\textbf{Reconstruction}. 
The final SR output is obtained by feeding concatenated HR feature
maps for all frames into a reconstruction module, similarly
to~\cite{haris2018deep}: $\texttt{SR}_t~=~f_{rec}([H_{t-1}, H_{t-2},
..., H_{t-n}])$. In our experiments, $f_{rec}$ is a single
convolutional layer.

\subsection{Multiple Projection}
\begin{figure}[!t]
\centering
\includegraphics[width=7cm]{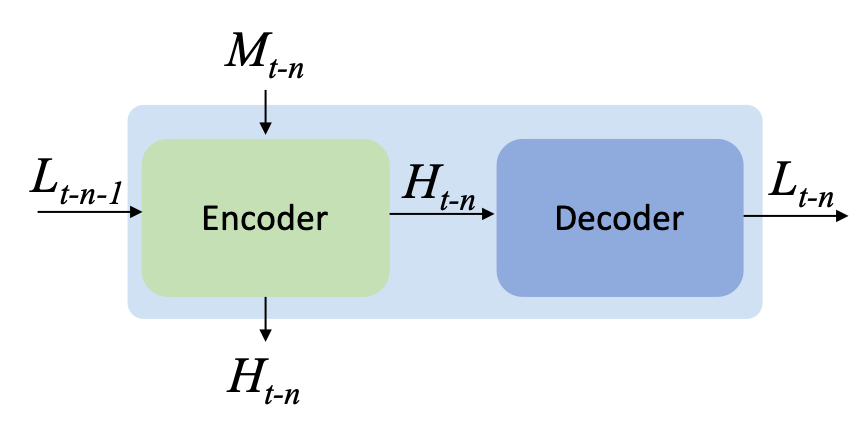}
\caption{The proposed projection module. The target features ($L_{t-n-1}$) is projected to neighbor features ($M_{t-n}$) to construct better HR features ($H_{t-n}$) and produce next LR features ($L_{t-n}$) for the next step.}
\label{figure:projection}
\end{figure}

The multiple projection stage of RBPN uses a recurrent chain of
encoder-decoder modules, as shown in Fig.~\ref{figure:projection}. 
The projection module, shared across time frames, takes two inputs: $L_{t-n-1}~\in~\mathbb{R}^{M^l \times N^l \times c^l}$ and $M_{t-n}~\in~\mathbb{R}^{M^l \times N^l \times c^m}$, then produces two outputs: $L_{t-n}$ and $H_{t-n}~\in~\mathbb{R}^{M^h \times N^h \times c^h}$ where $c^l, c^m, c^h$ are the number of channels for particular map accordingly.

\begin{figure}[!t]
  \begin{center}
    \begin{tabular}[c]{cc}
      \includegraphics[width=.4\textwidth]{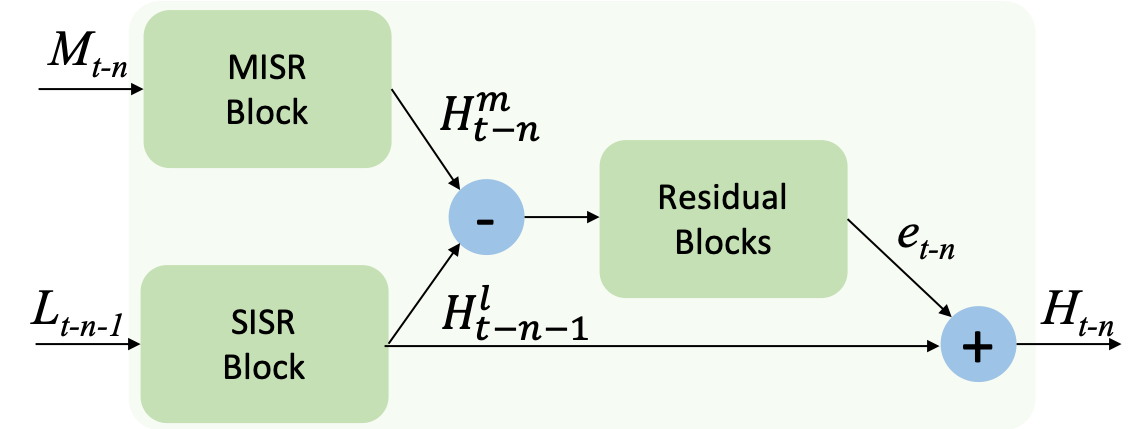}\\
      {\small (a) Encoder (the back-projection)\vspace{.2em}}\\
        \includegraphics[width=.2\textwidth]{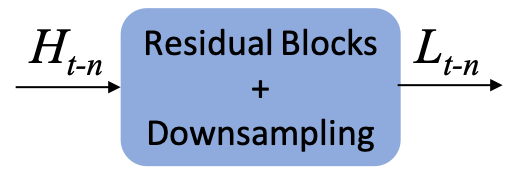}\\
      {\small (b) Decoder\vspace{.2em}}\\
    \end{tabular}
    \caption{Detailed illustration of encoder and decoder. The encoder performs back-projection from $L_{t-n-1}$ to $M_{t-n}$ to produce the residual $e_{t-n}$.}
    \label{fig:figure2}
  \end{center}\vspace{-1.5em}
\end{figure}

The encoder produces a hidden state of estimated HR features from the projection to a particular neighbor frame. 
The decoder deciphers the respective hidden state as the next input for the encoder module as shown in Fig.~\ref{fig:figure2} which are defined as follows:
\vspace{-1.5em}

\begin{align}\label{eq:networks}
&\text{Encoder:}&H_{t-n} &= \texttt{Net}_{E}(L_{t-n-1}, M_{t-n}; \theta_{E}) \\
&\text{Decoder:}&L_{t-n} &= \texttt{Net}_{D}(H_{t-n}; \theta_{D}) 
\end{align}

The encoder module $\texttt{Net}_E$ is defined as follows:
\begin{align}\label{eq:encoder}
&\text{SISR upscale:}~~H_{t-n-1}^l 	= \texttt{Net}_{sisr}(L_{t-n-1}; \theta_{sisr}) \\
&\text{MISR upscale:}~~H_{t-n}^m 	= \texttt{Net}_{misr}(M_{t-n}; \theta_{misr}) \\
&\text{Residual:}~~e_{t-n} 		=  \texttt{Net}_{res}(H_{t-n-1}^l  - H_{t-n}^m; \theta_{res}) \\
&\text{Output:}~~H_{t-n} 		= H_{t-n-1}^l + e_{t-n} 
\end{align}

\subsection{Interpretation}
\begin{figure*}[!t]
\centering
\includegraphics[width=15cm]{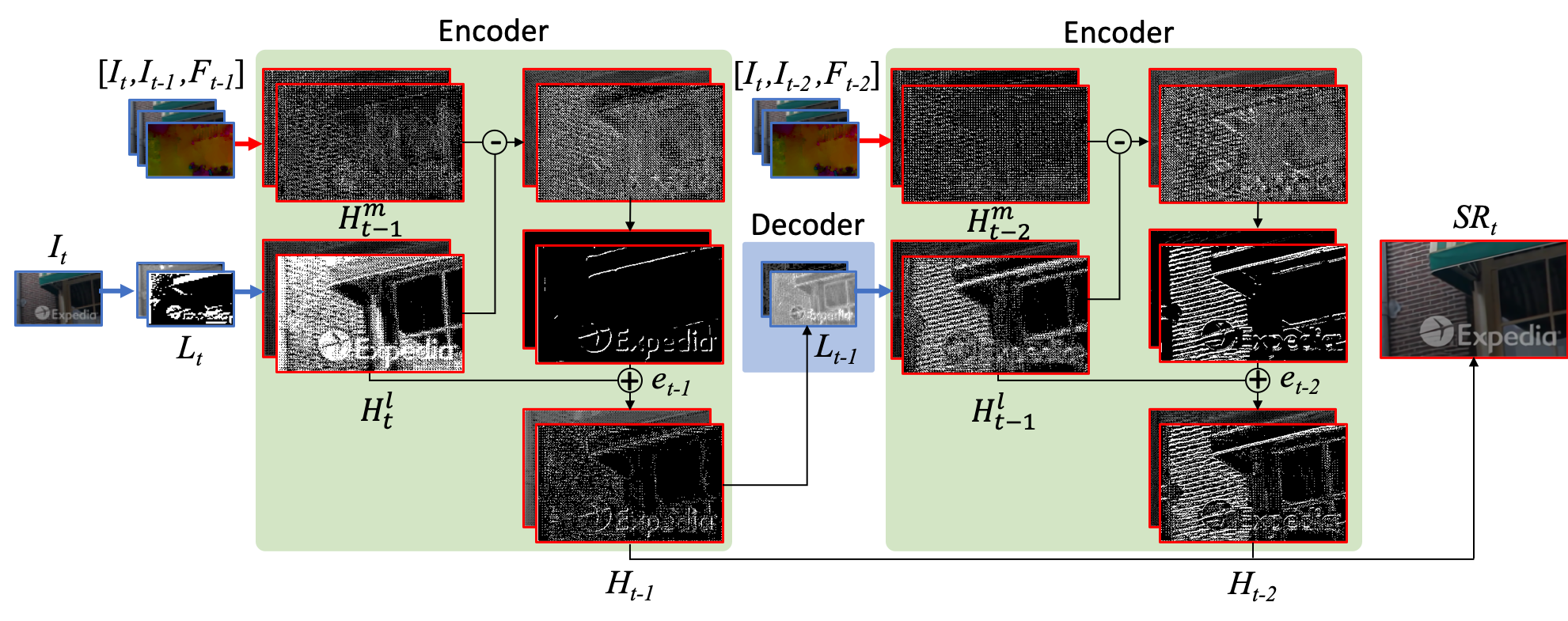}\vspace{-0.8em}
\caption{The illustration of each operation in RBPN ($n+1 = 3$). Zoom in to see better visualization.}
\label{figure:interpretation}
\end{figure*}

Figure~\ref{figure:interpretation} illustrates the RBPN pipeline, for
a 3-frame video.
In the encoder, we can see RBPN as the combination of SISR and MISR networks. First, target frame is enlarged by $\texttt{Net}_{sisr}$ to produce  $H_{t-k-1}^l$.
Then, for each combination of concatenation from neighbor frames and
target frame, $\texttt{Net}_{misr}$ performs implicit frame alignment
and absorbs the motion from neighbor frames to produce warping
features $H_{t-k}^m$ which may capture missing details in the target frame. 
Finally, the residual features $e_{t-k}$ from $H_{t-k-1}^l$ and $H_{t-k}^m$ are fused back to $H_{t-k-1}^l$ to refine the HR features and produce hidden state $H_{t-k}$. 
The decoder ``deciphers'' the hidden state $H_{t-k}$ to be the next input for the encoder $L_{t-k}$.
This process is repeated iteratively until the target frame is projected to all neighbor frames.

The optimal scenario for this architecture is when each frame can
contribute to filling in some missing details in the target
frame. Then $H_{t-k}$ generated in each step $k$ produce unique
features. In the generate case when $n=0$ (no other frames) or the
video is completely static (identical frames) RBPN will effectively
ignore the $\texttt{Net}_{misr}$ module, and fall back to a recurrent
SISR operation.

%% file: experiments.tex
\section{Experimental Results}
\label{sec:experiments}

In all our experiments, we focus on $4\times$ SR factor.
\subsection{Implementation and training details} 
We use DBPN~\cite{haris2018deep} for $\texttt{Net}_{sisr}$, and Resnet~\cite{he2015deep} for $\texttt{Net}_{misr}$, $\texttt{Net}_{res}$, and $\texttt{Net}_{D}$.
For $\texttt{Net}_{sisr}$, we construct three stages using $8 \times
8$ kernel with stride = 4 and pad by 2 pixels.
For $\texttt{Net}_{misr}$, $\texttt{Net}_{res}$, and $\texttt{Net}_{D}$, we construct five blocks where each block consists of two convolutional layers with $3 \times 3$ kernel with stride = 1 and pad by 1 pixel.
The up-sampling layer in $\texttt{Net}_{misr}$ and down-sampling layer in $\texttt{Net}_{D}$ use $8 \times 8$ kernel with stride = 4 and pad by 2 pixels.
Our final network uses $c^l = 256, c^m = 256$, and $c^h = 64$.

We trained our networks using Vimeo-90k~\cite{xue2017video}, with a
training set of 64,612 7-frame sequences, with fixed resolution 448 $\times$ 256.
Furthermore, we also apply augmentation, such as rotation, flipping,
and random cropping. To produce LR images, we downscale the HR images
$4\times$ with bicubic interpolation. 

All modules are trained end-to-end using per-pixel L1 loss per-pixel
between the predicted frame and the ground truth HR frame.
We use batch size of 8 with size $64 \times 64$ which is cropped randomly from $112\times 64$ LR image. 
The learning rate is initialized to $1e-4$ for all layers and decrease by a factor of 10
for half of total 150 epochs. We initialize the weights based
on~\cite{he2015delving}. For optimization, we used Adam with momentum
to $0.9$. All experiments were conducted
using Python 3.5.2 and PyTorch 1.0 on NVIDIA TITAN X GPUs. 
Following the evaluation from previous approaches~\cite{caballero2017real,tao2017detail, sajjadi2018frame},
we crop $8$ pixels near image boundary and
remove first six frames and last three frames. 
All measurements use only the luminance channel (Y).

\subsection{Ablation studies} 
\noindent{\textbf{Baselines}}
We consider three baselines, that retain some components of RBPN while
removing others. First, we remove all components by
$\texttt{Net}_{sisr}$ (DBPN); this ignores the video context. 
Second, we use DBPN with temporal concatenation (DBPN-MISR).
Third, we remove the decoder, thus severing temporal connections, so that our
model is reduced to applying back-projection $\texttt{Net}_{misr}$ with each
neighboring frame, and concatenating the results; we call this
baseline RBPN-MISR.
The results are shown in Table~\ref{tab:baseline}. 
Our intuition suggests, and the results confirm, that such an approach
would be weaker than RBPN, since it
does not have the ability to separately handle changes of different
magnitude that RBPN has. 
As expected,
SISR suffers from ignoring extra information in other
frames. RBPN-MISR and DBPN-MISR does manage to leverage multiple frames to improve
performance, but the best results are obtained by
the full RBPN model. 

\begin{table}[t!]
\scriptsize
\begin{center}
\begin{tabular}{*1c|*1c|*1c|*1c|*1c}
\hline
Bicubic&DBPN& DBPN-MISR& RBPN-MISR & RBPN \\     
1 Frame&1 Frame& 5 Frames & 5 Frames & 5 Frames \\     
\hline
27.13/0.749&29.85/0.837 & 30.64/0.859 &{30.89/0.866} & {31.40/0.877}\\
\hline
\end{tabular}
\end{center}
\caption{Baseline comparison on SPMCS-32. {\color{red}Red} here and in
  the other tables indicates the best performance (PSNR/SSIM).}
\label{tab:baseline}
\end{table}

\noindent{\textbf{Network setup.}} 
The modular design of our approach allows easy replacement of
modules; in particular we consider choices of DBPN or ResNet for
$\texttt{Net}_{sisr}$, $\texttt{Net}_{misr}$, or both. In
Table~\ref{tab:net}, we evaluate three combinations: RBPN with DBPN,
RBPN with Resnet, and RBPN with the combination of DBPN as
$\texttt{Net}_{sisr}$ and Resnet as $\texttt{Net}_{misr}$. The latter produces the best results, but the difference are minor,
showing stability of RBPN w.r.t. choice of components.

\begin{table}[t!]
\scriptsize
\begin{center}
\begin{tabular}{*1c|*1c|*1c}
\hline
DBPN & RESNET &$\texttt{Net}_{sisr}$=DBPN, $\texttt{Net}_{misr}$=RESNET \\     
\hline
30.54/0.856&30.74/0.862&{\color{red}30.96/0.866}\\
\hline
\end{tabular}
\end{center}
\caption{Network analysis using RBPN/2 on SPMCS-32. 
  }
\label{tab:net}
\end{table}

\begin{figure}[t!]
  \begin{center}
\includegraphics[width=8.5cm]{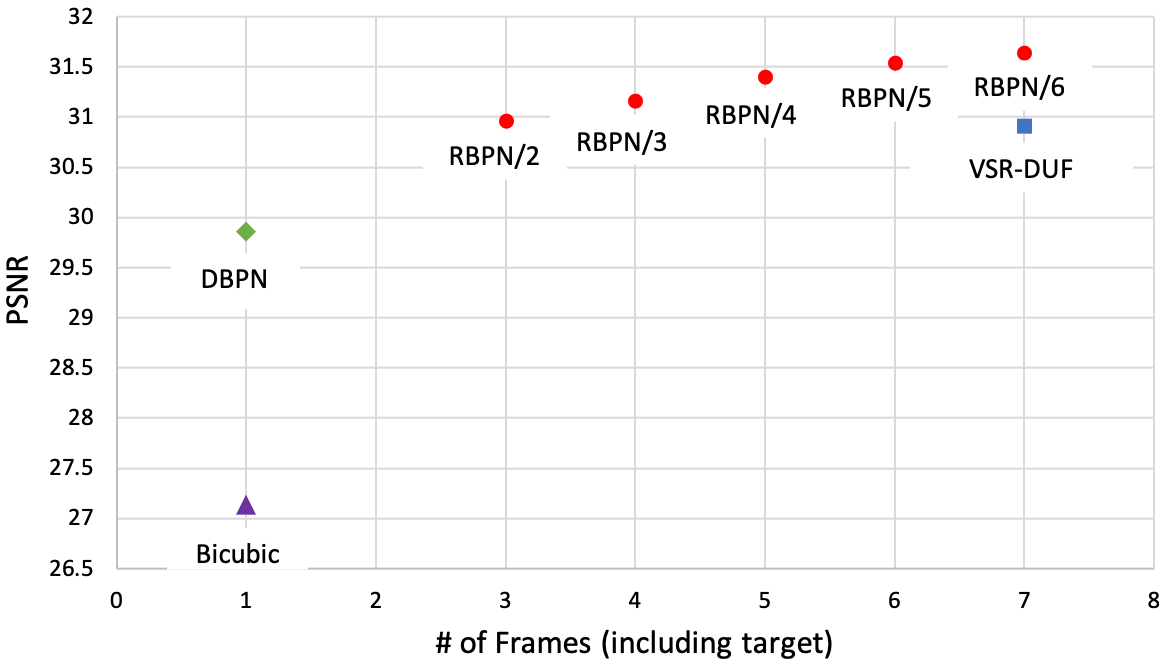}
\caption{Effect of context (past) length, $4\times$ SR on
  SPMCS-32. RBPN/$\langle k\rangle$: RBPN trained/tested with $k$
  past frames. Note: DBPN is equivalent to RBPN/0.}\vspace{-1em}
\label{figure:frames}
  \end{center}
\end{figure}

\noindent{\textbf{Context length}}
We evaluated RBPN with different lengths of video context, i.e.,
different number of past frames $n\in\{2,\ldots,6\}$.
Figure~\ref{figure:frames} shows that performance (measured on (on SPMCS-32 test set) improves with longer
context.
The performance of RBPN/3 is even better than VSR-DUF as one of state-of-the-art VSR which uses six neighbor frames.
It also shows that by adding more frames, the performance of RBPN increase by roughly $0.2$ dB.

\begin{figure*}[!t]
  \begin{center}
    \begin{tabular}[c]{ccccccc}
      \includegraphics[width=.135\textwidth]{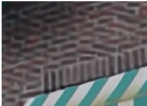}\hspace{-1em}
      &
      \includegraphics[width=.135\textwidth]{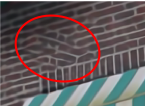}\hspace{-1em}
      &
       \includegraphics[width=.135\textwidth]{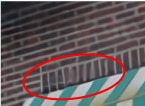}\hspace{-1em}
      &
       \includegraphics[width=.135\textwidth]{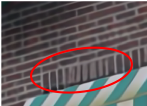}\hspace{-1em}
       &
       \includegraphics[width=.135\textwidth]{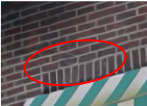}\hspace{-1em}
      &
        \includegraphics[width=.135\textwidth]{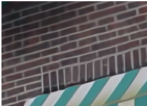}\hspace{-1em}
        &
        \includegraphics[width=.135\textwidth]{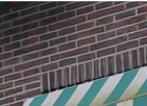}\hspace{-1em}\\
      {\small (a) VSR-DUF/6~\cite{jo2018deep}\hspace{-1em}}
      &{\small (b) RBPN/2\hspace{-1em}}
      &{\small (c) RBPN/3\hspace{-1em}}
      &{\small (d) RBPN/4\hspace{-1em}}
      &{\small (e) RBPN/5\hspace{-1em}}
      &{\small (f) RBPN/6\hspace{-1em}}
      &{\small (g) GT \vspace{0.5em}}\\
    \end{tabular}
    \caption{Visual results on different frame length (SPMCS-32). Zoom
      in to see better visualization.}\vspace{-1.5em}
    \label{fig:frames}
  \end{center}
\end{figure*}

Fig.~\ref{fig:frames} provides an illustration of the underlying
performance gains. Here, VSR-DUF fails to reconstruct the brick
pattern, while RBPN/3 reconstructs it well, even with fewer frames in
the context; increasing context length leads to further improvements.


\noindent{\textbf{Temporal integration}} 
Once the initial feature extraction and the projection modules have
produced a sequence of HR feature maps $H_{t-k}$, $k=1,\ldots,n$, we
can use these maps in multiple ways to reconstruct the HR target. The
proposed DBPN concatenates the maps; We also consider an alternative
where only the
$H_{t-n}$ is fed to $\texttt{Net}_{rec}$ (referred to as Last).
Furthermore, instead of concatenating the maps, we can feed them to a convolutional
LSTM~\cite{xingjian2015convolutional}, the output of which is then fed
to $\texttt{Net}_{rec}$.
The results are shown in Table~\ref{tab:lstm}. Dropping the
concatenation and only using last feature map harms the performance
(albeit moderately). Replacing concatenation with an LSTM also reduces
the performance (while increasing computational cost). We conclude that
the RBPN design depicted in Fig.~\ref{figure:proposed_network} is
better than the alternatives.


\begin{table}[t!]
\scriptsize
  \begin{center}
\begin{tabular}{*1l|*1c|*1c||*1c|*1c}
\hline
 &\multicolumn{2}{c}{RBPN/2}&\multicolumn{2}{c}{RBPN/6}  \\   
& RBPN & Last & w/ LSTM & RBPN \\     
\hline
PSNR/SSIM &{\color{red}30.96/0.866}&30.89/0.864&31.46/0.880&{\color{red}31.64/0.883}\\
\hline
\end{tabular}
  \end{center}
\caption{Comparison of temporal integration strategies on SPMCS-32.
}
\label{tab:lstm}
\end{table}

\noindent\textbf{Temporal order} When selecting frames to serve as
context for a target frame $t$, we have a choice of how to choose and
order it: use only past
frames (P; for instance, with $n=6$, this means $I_{t},I_{t-1},\ldots,I_{t-6}$), use both past and future
(PF, $I_{t-3},\ldots,I_t,\ldots,I_{t+3}$), or consider the past frames in
random order (PR; we can do this since the motion flow is computed
independently for each context frame w.r.t. the target).
Table~\ref{tab:temporal} shows that PF is better than P by 0.1 dB;
presumably this is due to the increased, more symmetric representation
of motion occurring in frame $t$. Interestingly, when the network is
trained on PF, then tested on P (PF$\,\to\,$P), the performance is
decreased (-0.17dB), but when RBPN is trained on P then tested on PF
(P$\,\to\,$PF), the performance remains almost the same. 

\begin{table}[t!]
\scriptsize
  \begin{center}
\begin{tabular}{*1l|*1c|*1c|*1c|*1c}
\hline
& P & PF & P$\,\to\,$PF & PF$\,\to\,$P \\     
\hline
PSNR/SSIM &31.64/0.883&{\color{red}31.74/0.884}&31.66/{\color{red}0.884}&31.57/0.881\\
\hline
\end{tabular}
  \end{center}
\caption{Effect of temporal order of context, RBPN/6 on SPMCS-32.
}
\label{tab:temporal}
\end{table}

The results of comparing order P to random ordering PR are shown in Table~\ref{tab:order}.
Interestingly, RBPN performance is not significantly affected by the
choice of order. We attribute this robustness to the decision to
associate each context frame with the choice of order.

\begin{table}[t!]
\scriptsize
  \begin{center}
\begin{tabular}{*1l|*1c|*1c|*1c|*1c}
\hline
& P & PR & P$\,\to\,$PR & PR$\,\to\,$P \\     
\hline
PSNR/SSIM &{\color{red}31.40/0.877}&31.39/0.876&31.39/0.877&31.35/0.875\\
\hline
\end{tabular}
  \end{center}
\caption{Effect of temporal order (RBPN/4) on SPMCS-32.
}
\label{tab:order}
\end{table}

%

\noindent{\textbf{Optical flow}} 
Finally, we can remove the optical flow component of $M_{t-k}$,
feeding the projection modules only the concatenated frame pairs. 
As Table~\ref{tab:noflow} shows,
explicit optical flow representation is somewhat, but not substantially, beneficial. We compute the flow using an
implementation of~\cite{liu2009beyond}.

\begin{table}[t!]
\small
  \begin{center}
\begin{tabular}{*1l|*1c|*1c}
\hline
 &\multicolumn{2}{c}{RBPN/5}  \\   
& w/ & w/o \\     
\hline
PSNR/SSIM &{\color{red}31.54/0.881}&31.36/0.878\\
\hline
\end{tabular}
\end{center}
\caption{Optical flow (OF) importance on SPMCS-32.
}
\label{tab:noflow}\vspace{-1em}
\end{table}

%

\subsection{Comparison with the-state-of-the-arts}
We compare our network with eight state-of-the-art SR algorithms:
DBPN~\cite{haris2018deep}, BRCN~\cite{huang2015bidirectional},
VESPCN~\cite{caballero2017real}, $B_{123}+T$~\cite{liu2017robust},
VSR-TOFLOW~\cite{xue2017video}, DRDVSR~\cite{tao2017detail},
FRVSR~\cite{sajjadi2018frame}, and VSR-DUF~\cite{jo2018deep}. Note: only VSR-DUF and DBPN provide full testing code without restrictions, and most of the previous methods use different training sets. Other methods provide only the estimated SR frames. 
For RBPN, we use $n=6$ with PF (past+future) order, which achieves the
best results, denoted as RBPN/6-PF.

\begin{figure*}[!t]
  \begin{center}
    \begin{tabular}[c]{ccccccc}
      \includegraphics[width=.135\textwidth]{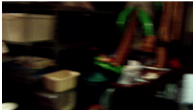}\hspace{-1em}
      &
      \includegraphics[width=.135\textwidth]{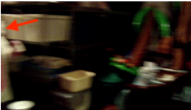}\hspace{-1em}
      &
      \includegraphics[width=.135\textwidth]{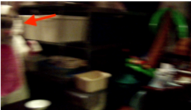}\hspace{-1em}
      &
      \includegraphics[width=.135\textwidth]{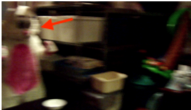}\hspace{-1em}
       &
        \includegraphics[width=.135\textwidth]{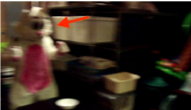}\hspace{-1em}
        &
       \includegraphics[width=.135\textwidth]{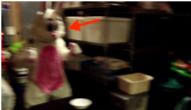}\hspace{-1em}
        &
        \includegraphics[width=.135\textwidth]{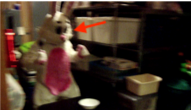}\hspace{-1em}\\
        
        \includegraphics[width=.135\textwidth]{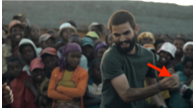}\hspace{-1em}
      &
      \includegraphics[width=.135\textwidth]{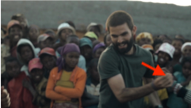}\hspace{-1em}
      &
      \includegraphics[width=.135\textwidth]{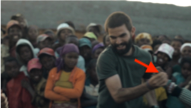}\hspace{-1em}
      &
      \includegraphics[width=.135\textwidth]{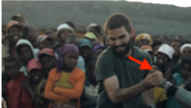}\hspace{-1em}
       &
        \includegraphics[width=.135\textwidth]{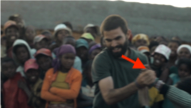}\hspace{-1em}
        &
       \includegraphics[width=.135\textwidth]{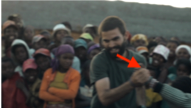}\hspace{-1em}
        &
        \includegraphics[width=.135\textwidth]{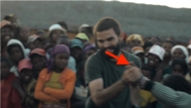}\hspace{-1em}\\
        
        \includegraphics[width=.135\textwidth]{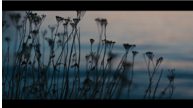}\hspace{-1em}
      &
      \includegraphics[width=.135\textwidth]{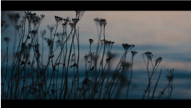}\hspace{-1em}
      &
      \includegraphics[width=.135\textwidth]{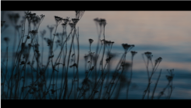}\hspace{-1em}
      &
      \includegraphics[width=.135\textwidth]{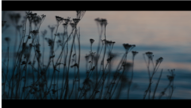}\hspace{-1em}
       &
        \includegraphics[width=.135\textwidth]{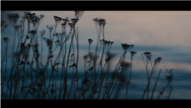}\hspace{-1em}
        &
       \includegraphics[width=.135\textwidth]{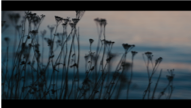}\hspace{-1em}
        &
        \includegraphics[width=.135\textwidth]{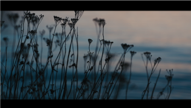}\hspace{-1em}\\
    \end{tabular}
    \caption{Examples from Vimeo-90k~\cite{xue2017video}. Top row:
      fast camera motion; new object appears in the third
      frame. Middle row: medium motion, little camer movement but some
      scene movement (e.g., person's arm in the foreground). Bottom
      row: slow motion only.}
    \label{figure:vimeo}
  \end{center}\vspace{-1em}
\end{figure*}

We carry out extensive experiments using three datasets: Vid4~\cite{liu2011bayesian}, SPMCS~\cite{tao2017detail}, and Vimeo-90k~\cite{xue2017video}. Each dataset has different characteristics. 
We found that evaluating on Vid4, commonly reported in literature, has
limited ability to assess relative merits of competing approaches; the
sequences in this set have visual artifacts, very little inter-frame
variation, and fairly limited motion. Most notably, it only consists
of \emph{four} video sequences.
SPMCS data exhibit more variation, but still lack significant
motion. Therefore, in addition to the aforementioned data sets, we
consider Vimeo-90k, a much larger and diverse data set, with
high-quality frames, and a range of motion types. We
stratify the Vimeo-90k sequences according to estimated motion
velocities into slow, medium and fast ``tiers'', as shown in
Fig.~\ref{figure:vimeo}, and report results for these tiers separately.

\begin{table*}[!t]
\scriptsize
\begin{center}
\scalebox{0.89}{
\begin{tabular}{*1l|*1c||*1c|*1c||*1c|*1c|*1c|*1c|*1c|*1c|*1c}
\hline
 & Flow & Bicubic & DBPN & BRCN &VESPCN&$B_{123}+T$&DRDVSR&FRVSR&VSR-DUF&RBPN/6-PF \\     
Clip Name & Magnitude &  & \cite{haris2018deep} & \cite{huang2015bidirectional} &\cite{caballero2017real}&\cite{liu2017robust}&\cite{tao2017detail}&\cite{sajjadi2018frame}&\cite{jo2018deep}& \\     
\hline
Calendar 		& 1.14 & 19.82/0.554 & 22.19/0.714 & - & - & 21.66/0.704 & 22.18/0.746 & - &({\color{red}24.09}/{\color{red}0.813}*)&{\color{blue}23.99/0.807} ({\color{blue}23.93}/{\color{blue}0.803}*) \\ 
City 			& 1.63 & 24.93/0.586 & 26.01/0.684 & - & - & 26.45/0.720 & 26.98/0.755 & - &({\color{red}28.26}/{\color{red}0.833}*)&{\color{blue}27.73/0.803} ({\color{blue}27.64}/{\color{blue}0.802}*) \\ 
Foliage 		& 1.48 & 23.42/0.575 & 24.67/0.662 & - & - & 24.98/0.698 & 25.42/0.720 & - &({\color{red}26.38}/{\color{red}0.771}*)&{\color{blue}26.22/0.757} ({\color{blue}26.27}/{\color{blue}0.757}*) \\ 
Walk 		& 1.44 & 26.03/0.802 & 28.61/0.870 & - & - & 28.26/0.859 & 28.92/0.875 & - &({\color{blue}30.50}/{\color{red}0.912}*)&{\color{red}30.70}/{\color{blue}0.909} ({\color{red}30.65}/{\color{blue}0.911}*) \\ 
\hline
Average 		& 1.42 & 23.53/0.629 & 25.37/0.737 & 24.43/0.662 & 25.35/0.756 & 25.34/0.745 & 25.88/0.774 & 26.69/{\color{blue}0.822} &({\color{red}27.31/0.832}*)&{\color{blue}27.12}/0.818 ({\color{blue}27.16}/0.819*) \\ 
\hline
\end{tabular}}
\end{center}
\caption{Quantitative evaluation of state-of-the-art SR algorithms on Vid4 for $4\times$. {\color{red}Red} indicates the best and {\color{blue}blue} indicates the second best performance (PSNR/SSIM). The calculation is computed without crop any pixels border and remove first and last two frames. For $B_{123}+T$ and DRDVSR, we use results provided by the authors on their webpage. For BRCN, VESPCN, and FRVSR, the values taken from their publications. *The output is cropped 8-pixels near image boundary.}
\label{tab:vid4}
\end{table*}

\begin{figure*}[t!]
  \begin{center}
    \begin{tabular}[c]{cccccc}
%
        
         \includegraphics[width=.155\textwidth]{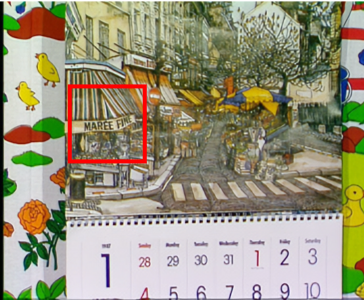}\hspace{-1em}\vspace{-0.3em}
      &
      \includegraphics[width=.15\textwidth]{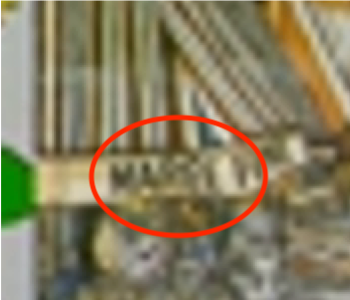}\hspace{-1em}
      &
      \includegraphics[width=.15\textwidth]{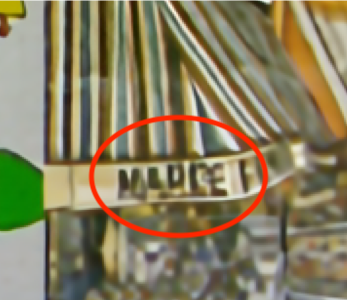}\hspace{-1em}
      &
      \includegraphics[width=.15\textwidth]{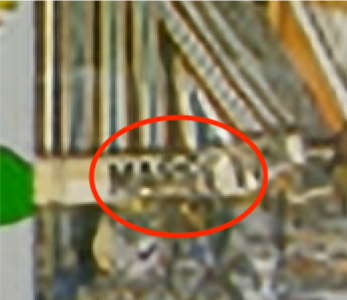}\hspace{-1em}
       &
       \includegraphics[width=.15\textwidth]{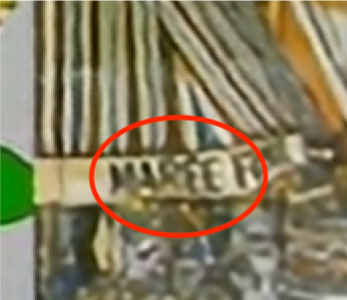}\hspace{-1em}
        &
        \includegraphics[width=.15\textwidth]{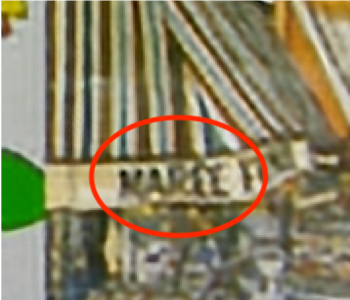}\hspace{-1em}\\
       ``Calendar''&{\small (a) Bicubic\hspace{-1em}}
      &{\small (b) DBPN~\cite{haris2018deep}\hspace{-1em}}
      &{\small (c) VSR~\cite{kappeler2016video}\hspace{-1em}}
      &{\small (d) VESPCN~\cite{caballero2017real}\hspace{-1em}}
      &{\small (e) $B_{123}+T$~\cite{liu2017robust} \vspace{0.1em}}\\
        
      &
      \includegraphics[width=.15\textwidth]{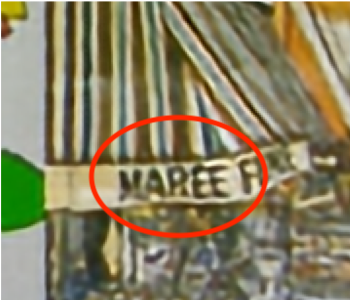}\hspace{-1em}\vspace{-0.3em}
      &
      \includegraphics[width=.15\textwidth]{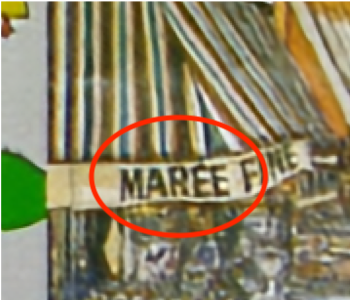}\hspace{-1em}
      &
      \includegraphics[width=.15\textwidth]{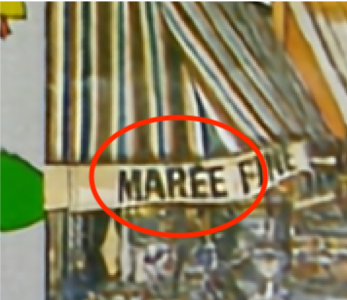}\hspace{-1em}
       &
       \includegraphics[width=.15\textwidth]{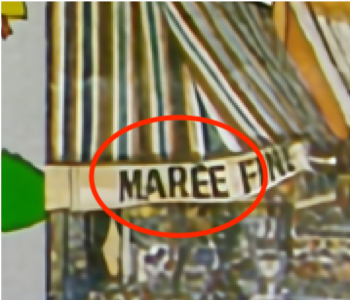}\hspace{-1em}
        &
        \includegraphics[width=.15\textwidth]{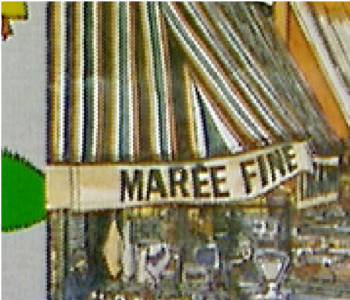}\hspace{-1em}\\
        &{\small (f) DRDVSR~\cite{tao2017detail}\hspace{-1em}}
      &{\small (g) FRVSR~\cite{sajjadi2018frame}\hspace{-1em}}
      &{\small (h) VSR-DUF~\cite{jo2018deep}\hspace{-1em}}
      &{\small (i) RBPN/6-PF\hspace{-1em}}
      &{\small (j) GT \vspace{0.2em}}\\
        
%
%
    \end{tabular}
    \caption{Visual results on Vid4 for $4\times$ scaling factor. Zoom in to see better visualization.}
    \label{fig:result_vid4}
  \end{center}\vspace{-1em}
\end{figure*}

Table~\ref{tab:vid4} shows the results on Vid4 test set. We also
provide the average flow magnitude (pixel/frame) on Vid4. It shows that Vid4 does not contain significant motion.
The results also show that RBPN/6-PF is better than the previous
methods, except for VSR-DUF.
Figure~\ref{fig:result_vid4} shows some qualitative results on Vid4.
(on ``Calendar''). The ``MAREE'' text reconstructed with RBPN/6-PF has
sharper images than previous methods. However, here we see that the
ground truth (GT) itself suffers from artifacts and aliasing, perhaps due to
JPEG compression. This apparently leads in some cases to penalizing
sharper SR predictions, like those made by our network, as illustrated
in Fig.~\ref{fig:result_vid4}.

\begin{table*}[!tb]
\scriptsize
\begin{center}
\begin{tabular}{*1l|*1c||*1c|*1c||*1c|*1c|*1c|*1c|*1c}
\hline
 & Flow & Bicubic & DBPN~\cite{haris2018deep} &DRDVSR~\cite{tao2017detail}&VSR-DUF~\cite{jo2018deep}&RBPN/4-P&RBPN/6-P&RBPN/6-PF \\     
Clip Name & Magnitude &  & (1 Frame) &(7 Frames)&(7 Frames)& (5 Frames)& (7 Frames)& (7 Frames) \\     
\hline
car05$\_$001 			& 6.21 & 27.62 & 29.58 & {\color{red}32.07} & 30.77 & 31.51 &31.65 & {\color{blue}31.92}   \\
hdclub$\_$003$\_$001	& 0.70 & 19.38 & 20.22 & 21.03 & {\color{red}22.07} &21.62& {\color{blue}21.91} & 21.88   \\
hitachi$\_$isee5$\_$001	& 3.01 & 19.59 & 23.47 & 23.83 & 25.73 &25.80& {\color{blue}26.14} & {\color{red}26.40}   \\
hk004$\_$001			& 0.49 & 28.46 & 31.59 & 32.14 & 32.96 &32.99& {\color{blue}33.25} &{\color{red} 33.31}   \\
HKVTG$\_$004		& 0.11 & 27.37 & 28.67 & 28.71 & 29.15 &29.28& {\color{blue}29.39} & {\color{red}29.43}   \\
jvc$\_$009$\_$001		& 1.24 & 25.31 & 27.89 & 28.15 & 29.26 &29.81& {\color{blue}30.17} & {\color{red}30.26}   \\
NYVTG$\_$006		& 0.10 & 28.46 & 30.13 & 31.46 & 32.29 &32.83& {\color{blue}33.09} & {\color{red}33.25}   \\
PRVTG$\_$012		& 0.12 & 25.54 & 26.36 & 26.95 & 27.47 &27.33& {\color{blue}27.52} & {\color{red}27.60}   \\
RMVTG$\_$011		& 0.18 & 24.00 & 25.77 & 26.49 & 27.63 &27.33& {\color{blue}27.64} & {\color{red}27.69}   \\
veni3$\_$011			& 0.36 & 29.32 & 34.54 & 34.66 & 34.51 &36.28& {\color{blue}36.14} & {\color{red}36.53}   \\
veni5$\_$015			& 0.36 & 27.30 & 30.89 & 31.51 & 31.75 &32.45& {\color{blue}32.66} & {\color{red}32.82}   \\
\hline
Average 				& 1.17 & 25.67/0.726 & 28.10/0.820 & 28.82/0.841 & 29.42/0.867 &29.75/0.866& {\color{blue}29.96/0.873} & {\color{red}30.10/0.874}  \\
\hline
\end{tabular}
\end{center}
\caption{Quantitative evaluation of state-of-the-art SR algorithms on SPMCS-11 for $4\times$. {\color{red}Red} indicates the best and {\color{blue}blue} indicates the second best performance (PSNR/SSIM).}
\label{tab:spmcs11}\vspace{-0.5em}
\end{table*}

\begin{table}[!tb]
\small
\begin{center}
\begin{tabular}{*1l|*1c|*1c|*1c}
\hline
 &\multicolumn{3}{c}{Vimeo-90k}  \\         
Algorithm  & Slow & Medium & Fast   \\
\hline
Bicubic &29.33/0.829 &31.28/0.867 &34.05/0.902\\
DBPN~\cite{haris2018deep} &32.98/0.901 &35.39/0.925 &37.46/0.944\\
\hline
\hline
TOFLOW~\cite{xue2017video} &32.16/0.889 &35.02/0.925 &37.64/0.942\\
VSR-DUF/6~\cite{jo2018deep} &32.96/0.909 &35.84/0.943 &37.49/0.949\\
RBPN/3-P &33.73/0.914 &36.66/0.941 &39.49/0.955\\
RBPN/6-PF &{\color{red}34.18/0.920} &{\color{red}37.28/0.947} &{\color{red}40.03/0.960}\\
\hline
\# of clips &1,616 &4,983 &1,225\\
Avg. Flow Mag. &0.6 &2.5 &8.3\\
\hline
\end{tabular}
\end{center}
\caption{Quantitative evaluation of state-of-the-art SR algorithms on
  Vimeo-90k~\cite{xue2017video} for $4\times$.
}
\label{tab:vimeo}\vspace{-1em}
\end{table}

\begin{figure*}[t!]
  \begin{center}
    \begin{tabular}[c]{cccccc}
      \includegraphics[width=.2258\textwidth]{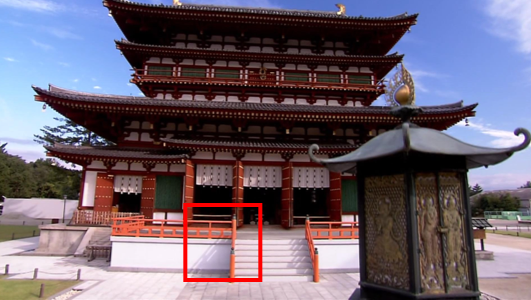}\hspace{-1em}
      &
      \includegraphics[width=.15\textwidth]{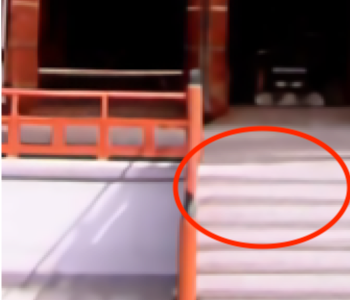}\hspace{-1em}
      &
      \includegraphics[width=.15\textwidth]{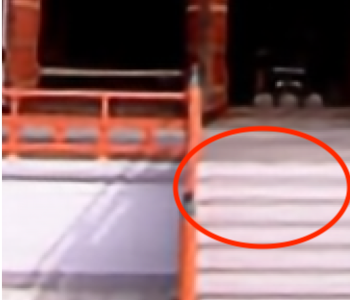}\hspace{-1em}
      &
      \includegraphics[width=.15\textwidth]{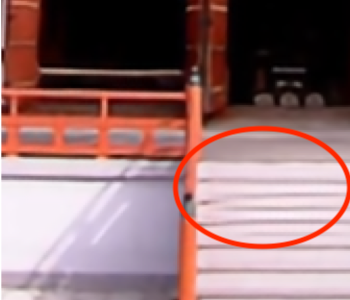}\hspace{-1em}
       &
       \includegraphics[width=.15\textwidth]{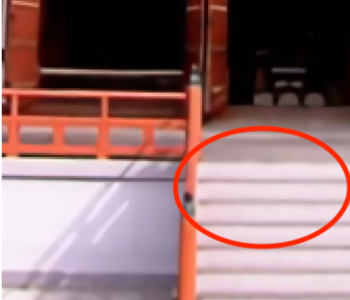}\hspace{-1em}
        &
        \includegraphics[width=.15\textwidth]{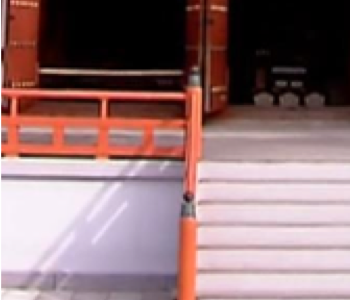}\hspace{-1em}\\
        
        \includegraphics[width=.2258\textwidth]{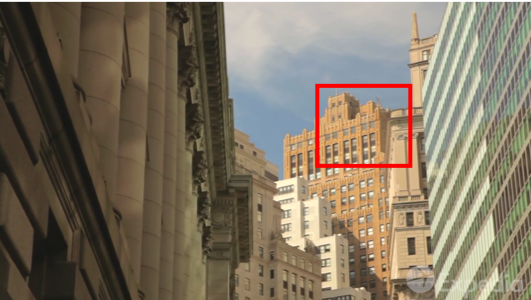}\hspace{-1em}
      &
      \includegraphics[width=.15\textwidth]{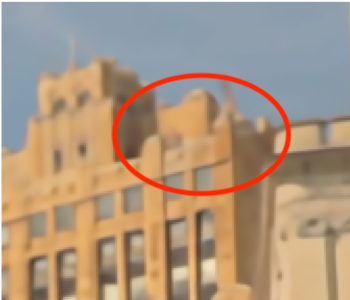}\hspace{-1em}
      &
      \includegraphics[width=.15\textwidth]{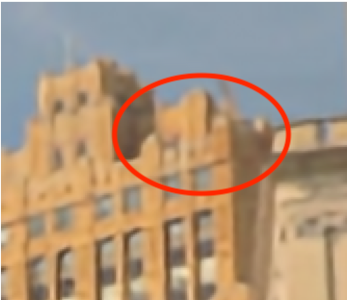}\hspace{-1em}
      &
      \includegraphics[width=.15\textwidth]{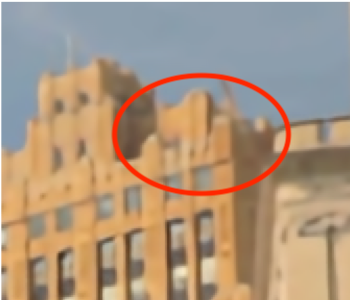}\hspace{-1em}
       &
       \includegraphics[width=.15\textwidth]{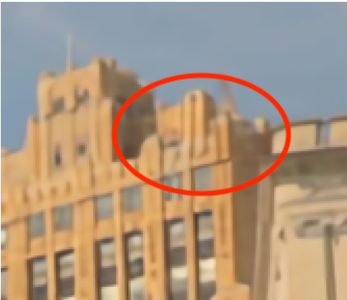}\hspace{-1em}
        &
        \includegraphics[width=.15\textwidth]{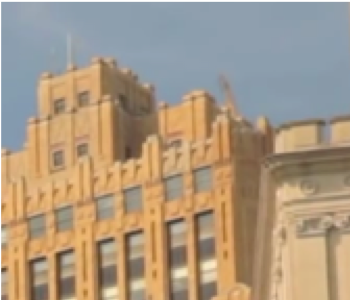}\hspace{-1em}\\
        
      &{\small (a) DBPN~\cite{haris2018deep}\hspace{-1em}}
      &{\small (b) DRDVSR~\cite{tao2017detail}\hspace{-1em}}
      &{\small (c) VSR-DUF~\cite{jo2018deep}\hspace{-1em}}
      &{\small (d) RBPN/6-PF\hspace{-1em}}
      &{\small (e) GT \vspace{0.2em}}\\
    \end{tabular}
    \caption{Visual results on SPMCS for $4\times$ scaling factor. Zoom in to see better visualization.}
    \label{fig:result_spmcs}
  \end{center}\vspace{-1em}
\end{figure*}

Table~\ref{tab:spmcs11} shows the detailed results on
SPMCS-11. RBPN/6-PF has better performance of 0.68 dB and 1.28 dB than
VSR-DUF and DRDVSR, respectively. Even with fewer frames in the context, RBPN/4-P has
better average performance than VSR-DUF and DRDVSR by 0.33 dB and 0.93
dB, respectively. Qualitative results on SPMCS are shown in
Fig.~\ref{fig:result_spmcs}. In the first row, we see that RBPN reproduces a well-defined pattern, especially on the stairs area. 
In the second row, RBPN recovers sharper details and produces better brown lines from the building pattern.

It is interesting to see that VSR-DUF tends to do better on SSIM
than on PSNR. It has been suggested that PSNR is more sensitive to
Gaussian noise, while SSIM is more sensitive to compression
artifacts~\cite{hore2010image}. VSR-DUF generates up-sampling filter
to enlarge the target frame. The use of up-sampling filter can keep
overall structure of target frame which tends to have higher
SSIM. However, since the residual image produced by VSR-DUF fails to
generate the missing details, PSNR tends to be lower. In contrast with
VSR-DUF, our focus is to fuse the missing details to the target
frame. However, if in some cases we generate sharper pattern than GT,
this causes lower SSIM. This phenomenon mainly can be observed in the Vid4 test set.

Table~\ref{tab:vimeo} shows the results on Vimeo-90k. 
RBPN/6-PF outperforms VSR-DUF by a large margin. RBPN/6-PF gets higher
PSNR by 1.22 dB, 1.44 dB, and 2.54 dB than VSR-DUF on, respectively,
slow, medium, and fast motion. It can be seen that RBPN is able to
preserve different temporal scale. RBPN achieves the highest gap
relative to prior work on fast motion. Even with reduced amount of
temporal context available, RBPN/3-P (using only 3 extra frames) does
better than previous methods like VSR-DUF using the full 6-extra frame
context. RBPN/3-P get higher PSNR by 0.77 dB, 0.82 dB, and 2 dB than VSR-DUF on slow, medium, and fast motion, respectively.



\begin{figure*}[t!]
  \begin{center}
    \begin{tabular}[c]{cccccc}
      \includegraphics[width=.15\textwidth]{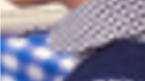}\hspace{-0.9em}
      &
      \includegraphics[width=.15\textwidth]{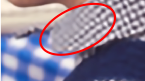}\hspace{-0.9em}
      &
      \includegraphics[width=.15\textwidth]{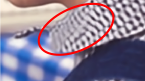}\hspace{-0.9em}
      &      
      \includegraphics[width=.15\textwidth]{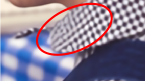}\hspace{-0.9em}
       &
       \includegraphics[width=.15\textwidth]{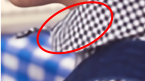}\hspace{-0.9em}
        &
        \includegraphics[width=.15\textwidth]{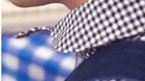}\hspace{-0.9em}\\
        
      \includegraphics[width=.15\textwidth]{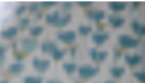}\hspace{-0.9em}
      &
      \includegraphics[width=.15\textwidth]{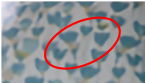}\hspace{-0.9em}
      &
      \includegraphics[width=.15\textwidth]{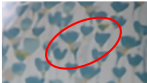}\hspace{-0.9em}
      &
      \includegraphics[width=.15\textwidth]{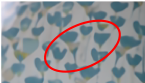}\hspace{-0.9em}
       &
       \includegraphics[width=.15\textwidth]{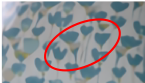}\hspace{-0.9em}
        &
        \includegraphics[width=.15\textwidth]{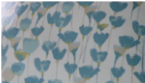}\hspace{-0.9em}\\

      \includegraphics[width=.15\textwidth]{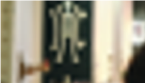}\hspace{-0.9em}
      &
      \includegraphics[width=.15\textwidth]{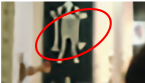}\hspace{-0.9em}
      &
      \includegraphics[width=.15\textwidth]{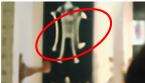}\hspace{-0.9em}
      &
      \includegraphics[width=.15\textwidth]{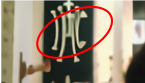}\hspace{-0.9em}
       &
       \includegraphics[width=.15\textwidth]{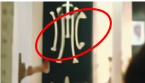}\hspace{-0.9em}
        &
        \includegraphics[width=.15\textwidth]{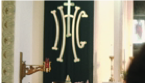}\hspace{-0.9em}\\

      \includegraphics[width=.15\textwidth]{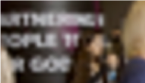}\hspace{-0.9em}
      &
      \includegraphics[width=.15\textwidth]{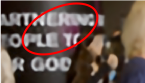}\hspace{-0.9em}
      &
      \includegraphics[width=.15\textwidth]{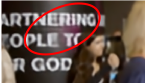}\hspace{-0.9em}
      &
      \includegraphics[width=.15\textwidth]{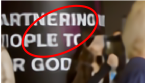}\hspace{-0.9em}
       &
       \includegraphics[width=.15\textwidth]{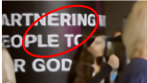}\hspace{-0.9em}
        &
        \includegraphics[width=.15\textwidth]{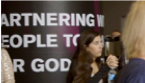}\hspace{-0.9em}\\

         {\small (a) Bicubic\hspace{-0.9em}}
      &{\small (b) TOFlow~\cite{xue2017video}\hspace{-0.9em}}
      &{\small (c) VSR-DUF~\cite{jo2018deep}\hspace{-0.9em}}
      &{\small (d) RBPN/3-P\hspace{-0.9em}}
      &{\small (e) RBPN/6-PF\hspace{-0.9em}}
      &{\small (f) GT \vspace{0.3em}}\\
    \end{tabular}
    \caption{Visual results on Vimeo-90k for $4\times$ scaling factor. Zoom in to see better visualization.}
    \label{fig:result_vimeo}
  \end{center}\vspace{-1.5em}
\end{figure*}

Figure~\ref{fig:result_vimeo} shows qualitative results on Vimeo-90k. 
RBPN/6-PF obtains reconstruction that appears most similar to the GT,
more pleasing and sharper than reconstructions with other methods. We
have highlighted regions in which this is particularly notable.

%% file: conclusion.tex
\section{Conclusion}
\label{conclusion}
We have proposed a novel approach to video super-resolution (VSR) called
Recurrent Back-Projection Network (RBPN). It's a modular architecture, in
which temporal and spatial information is collected from video frames
surrounding the target frame, combining ideas from single- and
multiple-frame super resolution. Temporal context is organized by a
recurrent process using the idea of (back)projection, yielding gradual
refinement of the high-resolution features used, eventually, to
reconstruct the high-resolution target frame. In addition to our
technical innovations, we propose a new evaluation protocol for video
SR. This protocol allows to differentiate performance of video SR
based on magnitude of motion in the input videos. In extensive
experiments, we assess the role played by various design choices in
the ultimate performance of our approach, and demonstrate that,
on a vast majority of thousands of test video sequences, RBPN obtains significantly better performance than existing VSR methods.